\documentclass[twocolumn,final]{svjour2}

\smartqed  

\usepackage[english]{babel}
\usepackage{amsmath}
\usepackage{amssymb}
\usepackage{amsfonts}
\usepackage{graphicx}
\usepackage{epstopdf}
\usepackage[justification=centering]{caption}
\usepackage{subcaption}
\usepackage{multirow}
\usepackage{booktabs}
\usepackage{lineno}
\usepackage{float}
\usepackage{setspace}
\usepackage{overpic}
\usepackage{rotating}
\usepackage[numbers]{natbib}
\usepackage{paralist}

\usepackage{algorithm}
\usepackage{algpseudocode}
\algnewcommand\algorithmicinput{\textbf{Input:}}
\algnewcommand\Input{\item[\algorithmicinput]}

\usepackage{hyperref}
\hypersetup{
	colorlinks,
	citecolor=blue,
	filecolor=blue,
	linkcolor=blue,
	urlcolor=blue
}

\DeclareMathOperator*{\argmin}{arg\rm{}min}

\newcommand{\bhs}{\mathbf{\hat{s}}}

\newcommand{\bY}{\mathbf{Y}}

\newcommand{\bbeta}{\boldsymbol{\beta}}

 \usepackage{tikz}
\usetikzlibrary{positioning,shapes,shadows,arrows,calc}

\tikzstyle{block} = [draw,line width=1.2pt, fill=white!30, rectangle, 
    minimum height=3em, minimum width=3em, rounded corners]         
\tikzstyle{blocky} = [draw, line width=1.2pt,fill=yellow!40, rectangle, 
    minimum height=2.5em, minimum width=2.5em, rounded corners]    
\tikzstyle{blockp} = [draw,line width=1.2pt, fill=purple!30, rectangle, 
    minimum height=3em, minimum width=3em, rounded corners]       
\tikzstyle{blockb} = [draw,line width=1.2pt, fill=blue!20, rectangle, 
    minimum height=2.5em, minimum width=2.5em, rounded corners]     
\tikzstyle{blockr} = [draw, line width=1.2pt,fill=red!30, rectangle, 
    minimum height=2.5em, minimum width=2.5em, rounded corners]                 
    \tikzstyle{blockgr} = [draw, line width=1.2pt,fill=black!10, rectangle, 
    minimum height=2.5em, minimum width=2.5em, rounded corners]                  
\tikzstyle{line}=[-, line width=1.2pt]
    \tikzstyle{blocky4} = [draw, line width=1.2pt,fill=red!35, rectangle, 
        text centered,minimum height=4.5em, text width=6.5em, rounded corners]        
    \tikzstyle{blockb4} = [draw, line width=1.2pt,fill=blue!20, rectangle, 
        text centered,minimum height=4.5em, text width=6.5em, rounded corners]    

\journalname{}

\begin{document}
\title{Compressed Dynamic Mode Decomposition for Background Modeling}
\titlerunning{Compressed DMD} 

\author{N. Benjamin Erichson\and Steven L. Brunton\and J. Nathan Kutz}

\institute{N. Benjamin Erichson \at
		   School of Mathematics and Statistics \\
	       University of St Andrews \\
	       St Andrews, United Kingdom\\
           \email{nbe@st-andrews.ac.uk}           
           \and
            Steven L. Brunton \at
            Department of Mechanical Engineering\\
            University of Washington\\
            Seattle, WA 98195
            \and
            J. Nathan Kutz \at
            Department of Applied Mathematics\\
            University of Washington\\
            Seattle, WA 98195-2420
}

\date{}

\maketitle
\sloppy

\begin{abstract}
	
	%
	We introduce the method of compressed dynamic mode decomposition (cDMD) for background modeling. 
	The dynamic mode decomposition (DMD) is a regression technique that integrates two of the leading data analysis methods in use today: Fourier transforms and singular value decomposition.  
	Borrowing ideas from compressed sensing and matrix sketching, cDMD eases the computational workload of high resolution video processing. The key principal of cDMD is to obtain the decomposition on a (small) compressed matrix representation of the video feed. 
	Hence, the cDMD algorithm scales with the intrinsic rank of the matrix, 
	rather then the size of the actual video (data) matrix. 	
	Selection of the optimal modes characterizing the background is formulated as a sparsity-constrained sparse coding problem.
	%
	%
	Our results show, that the quality of the resulting background model is competitive, quantified by the F-measure, Recall and Precision.  
	A GPU (graphics processing unit) accelerated implementation is also presented which further boosts the computational efficiency of the algorithm.  

\end{abstract}
\keywords{dynamic mode decomposition; background modeling; matrix sketching; sparse coding; GPU-accelerated computing.}
\vspace{+1.1in}


\section{Introduction}
\label{intro}

One of the fundamental computer vision objectives is to detect moving objects in a given video stream. At the most basic level, moving objects can be found in a video by removing the background. However, this is a challenging task in practice, since the true background is often unknown. Algorithms for background modeling are required to be both robust and adaptive. Indeed, the list of challenges is significant and includes camera jitter, illumination changes, shadows and dynamic backgrounds. There is no single method currently available that is capable of handling all the challenges in real-time without suffering performance failures. Moreover, one of the great challenges in this field is to efficiently process high-resolution video streams, a task that is at the edge of performance limits for state-of-the-art algorithms. Given the importance of background modeling, a variety of mathematical methods and algorithms have been developed over the past decade. Comprehensive overviews of traditional and state-of-the art methods are provided by Bouwmans \cite{bouwmans2014traditional} or Sobral and Vacavant \cite{Sobralreview}.

\paragraph{Motivation.} This work advocates the method of dynamic mode decomposition (DMD), which enables the decomposition of spatio-temporal grid data in both space and time. The DMD has been successfully applied to videos~\cite{grosek2014,erichson2015,mrDMDbg}, however the computational costs are dominated by the singular value decomposition (SVD). Even with the aid of recent innovations around randomized algorithms for computing the SVD~\cite{halko2011rand}, the computational costs remain expensive for high resolution videos. Importantly, we build on the recently introduced compressed dynamic mode decomposition (cDMD) algorithm, which integrates DMD with ideas from compressed sensing and matrix sketching~\cite{cdmd}. Hence, instead of computing the DMD on the full-resolution video data, we show that an accurate decomposition can be obtained from a compressed representation of the video in a fraction of the time. The optimal mode selection for background modeling is formulated as a sparsity-constrained sparse coding problem, which can be efficiently approximated using the greedy orthogonal matching pursuit method. The performance gains in computation time are significant, even competitive with Gaussian mixture-models. Moreover, the performance evaluation on real-videos shows that the detection accuracy is competitive compared to leading robust principal component analysis (RPCA) algorithms. 

\paragraph{Organization.} The rest of this paper is organized as follows. Section \ref{sec:video} presents a brief introduction to the dynamic mode decomposition and its application to video and background modeling. Section \ref{sec:cdmd} presents the compressed DMD algorithm and different measurement matrices to construct the compressed video matrix. A GPU accelerated implementation is also outlined. Finally a detailed evaluation of the algorithm is presented in section \ref{sec:results}. Concluding remarks and further research directions are given in section \ref{sec:conclusion}. Appendix~\ref{app:notation} gives an overview of notation. 

\section{DMD for Video Processing}
\label{sec:video}
\subsection{The Dynamic Mode Decomposition}
\begin{figure*}[htp]
	\centering
	\includegraphics[width=0.9\textwidth] {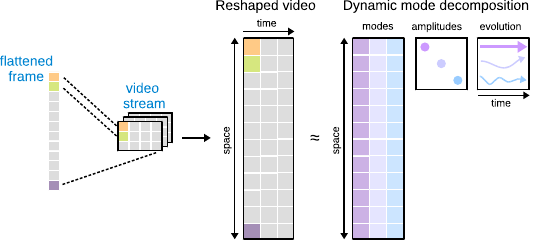}
	\caption{Illustration of the dynamic mode decomposition for video applications. Given a
		video stream, the first step involves reshaping the grayscale video frames into a 2-dimensional
		spatio-temporal grid. The DMD then creates a decomposition in space and time in which DMD modes contain spatial structure. }
	\label{fig:DMDvideo}
\end{figure*}

The dynamic mode decomposition is an equation-free, data-driven matrix decomposition that is capable of providing accurate reconstructions of spatio-temporal coherent structures arising in nonlinear dynamical systems, or short-time future estimates of such systems. DMD was originally introduced in the fluid mechanics community by Schmid~\cite{DMD1} and Rowley et al.~\cite{DMD4}. A surveillance video sequence offers an appropriate application for DMD because the frames of the video are, by nature, equally spaced in time, and the pixel data, collected in every snapshot, can readily be vectorized. The dynamic mode decomposition is illustrated for videos in Figure~\ref{fig:DMDvideo}. For computational convenience the flattened grayscale video frames (snapshots) of a given video stream are stored, ordered in time, as column vectors ${\bf x}_{1}, {\bf x}_{2}, \ldots, {\bf x}_{m}$ of a matrix. Hence, we obtain a 2-dimensional $\mathbb{R}^{n \times m}$ spatio-temporal grid, where $n$ denotes the number of pixels per frame, $m$ is the number of video frames taken, and the matrix elements $x_{it}$ correspond to a pixel intensity in space and time. The video frames can be thought of as snapshots of some underlying  dynamics. Each video frame (snapshot) $\mathbf{x}_{t+1}$ at time $t+1$ is assumed to be connected to the previous frame $\mathbf{x}_{t}$ by a linear map $\mathbf{A}: \mathbb{R}^{n}\rightarrow\mathbb{R}^{n}$. 
Mathematically, the linear map $\mathbf{A}$ is a time-independent operator which constructs the approximate linear evolution
\begin{equation}
\mathbf{x}_{t+1} = \mathbf{A}\mathbf{x}_{t}. 
\label{eq:dmdvec} 
\end{equation}
The objective of dynamic mode decomposition is to find an estimate for the matrix ${\mathbf{A}}$ and its eigenvalue decomposition that characterize the system dynamics. At its core, dynamic mode decomposition is a regression algorithm. First, the spatio-temporal grid is separated into two overlapping sets of data, called the left and right snapshot sequences
%
\begin{align}
{\bf X} \!=\! \begin{bmatrix}
\vline & \vline & & \vline \\
{\bf x}_1 & {\bf x}_2 & \cdots & {\bf x}_{m-1}\\
\vline & \vline & & \vline
\end{bmatrix}, \hspace{0.1in} {\bf X}' \!=\! \begin{bmatrix}
\vline & \vline & & \vline \\
{\bf x}_2 & {\bf x}_3 & \cdots & {\bf x}_m\\
\vline & \vline & & \vline
\end{bmatrix}.\label{Eq:FullData}
\end{align}
%
Equation \eqref{eq:dmdvec} is reformulated in matrix notation 
\begin{equation}
{\bf X}' = {\bf A}{\bf X}.
\end{equation}
In order to find an estimate for the matrix ${\bf A}$ we face the following least-squares problem
\begin{equation}
\hat{\bf A} =	\argmin_{\bf A} \|{\bf X}' - {\bf A}{\bf X}\|_{F}^{2},
\end{equation}
where $\|\cdot\|_{F}$ denotes the Frobenius norm. This is a well-studied problem, and an estimate of the linear operator ${\bf A}$ is given by
\begin{equation}
{\bf \hat{A}} = {\bf X}'{\bf X}^\dag,
\end{equation}
where $\dag$ denotes the Moore-Penrose pseudoinverse, which produces a regression that is optimal in a least-square sense. The DMD modes ${\bf \Phi=W}$, containing the spatial information, are then obtained as eigenvectors of the matrix ${\bf \hat{A}}$
\begin{equation}
{\bf \hat{A}}{\bf W} = {\bf W}\boldsymbol{\Lambda},
\end{equation}
where columns of ${\bf W}$ are eigenvectors ${\bf \phi}_j$ and $\boldsymbol{\Lambda}$ is a diagonal matrix containing the corresponding eigenvalues $\lambda_j$. 
In practice, when the dimension $n$ is large, the matrix ${\bf \hat{A}}\in \mathbb{R}^{n \times n}$ may be intractable to estimate and to analyze directly. DMD circumvents the computation of ${\bf \hat{A}}$ by considering a rank-reduced representation $\tilde{\bf A} \in \mathbb{R}^{k \times k}$. This is achieved by using the similarity transform, i.e., projecting $\tilde{\bf A}$ on the left singular vectors. Moreover, the DMD typically makes use of low-rank structure so that the total number of modes, $k \leq min(n,m)$, allows for dimensionality reduction of the video stream. Hence, only the relatively small $\tilde{\bf A} \in \mathbb{R}^{k \times k}$ matrix needs to be estimated and analyzed (see Section \ref{sec:cdmd} for more details). 
The dynamic mode decomposition yields then the following low-rank factorization of a given spatio-temporal grid (video stream):
\begin{equation}
{\bf \Phi}{\bf B}{\mathcal{V}} =
\scalebox{0.68}{ 
	$\begin{pmatrix}
	\phi_{11} & \phi_{1p} & \cdots & \phi_{1k}	\\
	\vdots  & \vdots & \ddots & \vdots  		\\	
	\phi_{i1} & \phi_{ip} & \cdots & \phi_{ik} 	\\
	\vdots  & \vdots & \ddots & \vdots  		\\
	\phi_{n1} & \phi_{np} & \cdots & \phi_{nk}
	\end{pmatrix}
	\begin{pmatrix}
	b_{1} 	&       	&         	&  		&\\
	& \ddots    &  			&   	&\\
	&			& b_{p} 	&  			&  		\\
	&       	&			& \ddots 	&   	\\
	&  			&			&  			& b_{k}
	\end{pmatrix}
	\begin{pmatrix}
	1 		& \lambda_{1} & \cdots & \lambda_{1}^{m-1}  \\
	\vdots	& \vdots  & \ddots & \vdots  		\\
	1		& \lambda_{p} & \cdots & \lambda_{p}^{m-1} 	\\
	\vdots	& \vdots  & \ddots & \vdots  		\\
	1		& \lambda_{k} & \cdots & \lambda_{k}^{m-1}
	\end{pmatrix}$}
\end{equation} 
where the diagonal matrix ${\bf B} \in \mathbb{C}^{k \times k}$ has the amplitudes as entries and ${ \mathcal{V}} \in \mathbb{C}^{k \times m}$ is the Vandermonde matrix describing the temporal evolution of the DMD modes ${\bf \Phi} \in \mathbb{C}^{n \times k}$. 

\subsection{DMD for Foreground/Background Separation}
The DMD method can attempt to reconstruct any given frame, or even possibly future frames. The validity of the reconstruction thereby depends on how well the specific video sequence meets the assumptions and criteria of the DMD method. Specifically, a video frame ${\bf x}_t$ at time points $t \in 1,...,m$ is approximately reconstructed as follows
\begin{equation}
{\bf \tilde{x}}_t =  \sum_{j=1}^k b_j \phi_j \lambda_j^{t-1}.
\label{eq:omegaj}
\end{equation}
Notice that the DMD mode $\phi_{j}$ is a $n \times 1$ vector containing the spatial structure of the decomposition, while the eigenvalue $\lambda_j^{t-1}$ describes the temporal evolution. The scalar $b_j$ is the amplitude of the corresponding DMD mode. At time $t=1$, equation \eqref{eq:omegaj} reduces to ${\bf \tilde{x}}_1 = \sum_{j=1}^k b_j \phi_j$. Since the amplitude is time-independent, $b_j$ can be obtained by solving the following least-square problem using the video frame ${\bf x}_1$ as initial condition
\begin{equation}\label{eq:b}
{\bf \hat{b}} = \argmin_{\bf b} \|{\bf x}_1 - {\bf \Phi}{\bf b}\|_{F}^{2}.
\end{equation}  
It becomes apparent that any portion of the first video frame that does not change in time, or changes very slowly in time, must have an associated continuous-time eigenvalue 
\begin{equation}
\omega_{j}= \dfrac{\log(\lambda_j)}{\Delta t}
\end{equation} 
that is located near the origin in complex space: $|\omega_{j}| \approx 0$ or equivalent $|\lambda_{j}| \approx 1$.
%
\begin{figure*}[htp]
	\centering
	\begin{subfigure}[t]{0.8\textwidth}
		\centering
		\DeclareGraphicsExtensions{.pdf}
		\includegraphics[width=0.9\textwidth]{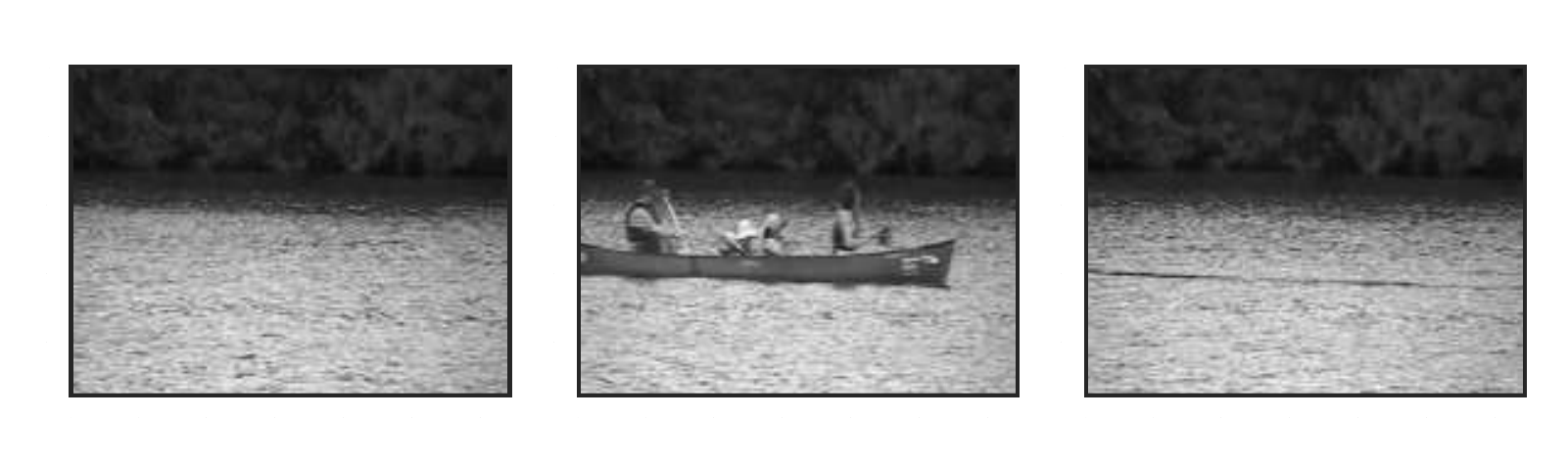}
		\caption{Sample frames ($t={0,150,300}$) of video sequence.}
	\end{subfigure}	

	\vspace{+0.1in}	
	\begin{subfigure}[t]{0.49\textwidth}
		\centering
		\DeclareGraphicsExtensions{.pdf}
		\includegraphics[width=1\textwidth]{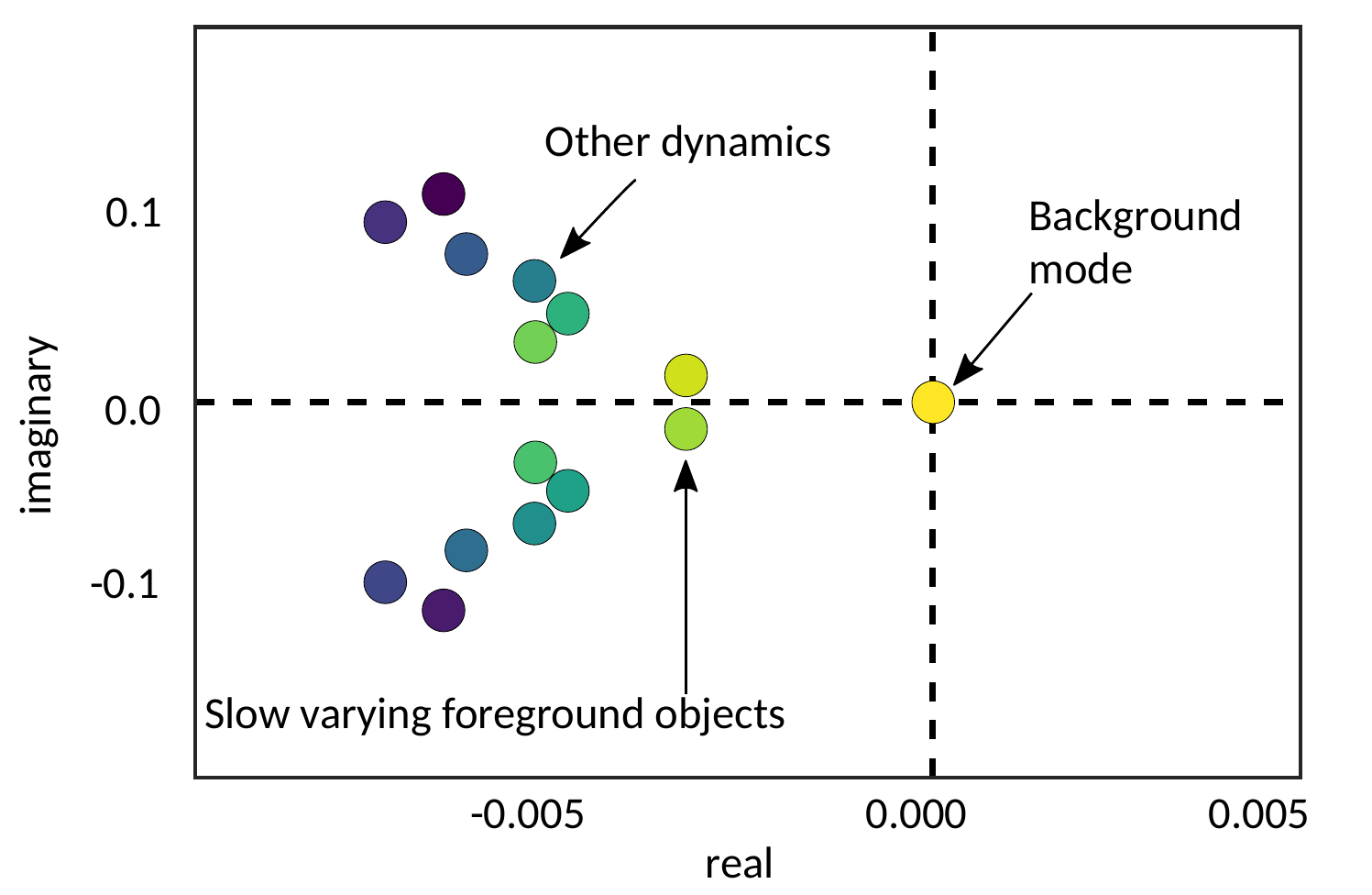}
		\caption{Dominant continuous-time eigenvalues $\omega_{j}$.}
	\end{subfigure}
	~
	\begin{subfigure}[t]{0.49\textwidth}
		\centering
		\DeclareGraphicsExtensions{.pdf}
		\includegraphics[width=1\textwidth]{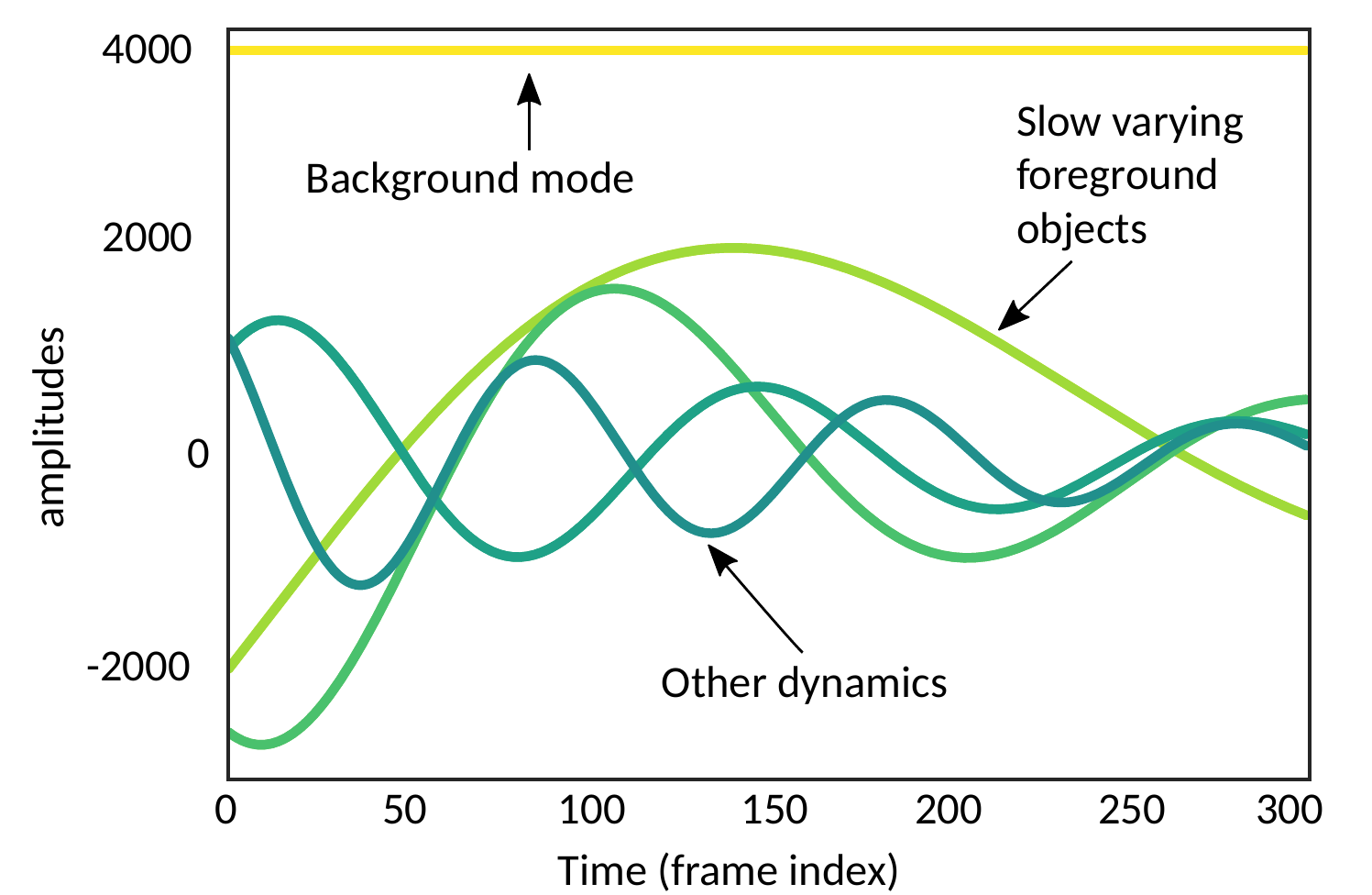}
		\caption{Amplitudes over time.}
	\end{subfigure}
	\caption{Results of the dynamic mode decomposition for the ChangeDetection.net video sequence `canoe'. Subplot (a) shows three samples frames of the video sequence. Subplot (b) and (c) show the the continuous-time eigenvalues and the temporal evolution of the amplitudes. The modes corresponding to the amplitudes with the highest variance are capturing the dominant foreground object (canoe), while the zero mode is capturing the dominant structure of the background. Modes corresponding to high frequency amplitudes capturing other dynamics in the video sequence, e.g., waves, etc.}	
	\label{fig:modes}
\end{figure*}
This fact becomes the key principle to separate foreground elements (approximate sparse) from background (approximate low-rank) information. Figure~\ref{fig:modes} shows the dominant continuous-time eigenvalues for a video sequence. Subplot (a) shows three sample frames from this video sequence that includes a canoe. Here the foreground object (canoe) is not present at the beginning and the end for the video sequence. The dynamic mode decomposition factorizes this sequence into modes describing the different dynamics present. The analysis of the continuous-time eigenvalue $\omega_{j}$ and the amplitudes over time ${\bf B\mathcal{V}}$ (the amplitudes multiplied by the Vandermonde matrix) can provide interesting insights, shown in subplot (b) and (c). First, the amplitude for the prominent zero mode (background) is constant over time, indicating that this mode is capturing the dominant (static) content of the video sequence, i.e, the background. The next pair of modes correspond to the canoe, a foreground object slowly moving over time. The amplitude reveals the presence of this object. Specifically, the amplitude reaches its maximum at about the frame index 150, when the canoe is in the center of the video frame. At the beginning and end of the video the canoe is not present, indicated by the negative values of the amplitude. The subsequent modes describe other dynamics in the video sequence e.g., the movements of the canoeist and the waves. For instance, the modes describing the waves have high frequency and small amplitudes (not shown here). Hence, a theoretical viewpoint we will build upon with the DMD methodology centers around the recent idea of low-rank and sparse matrix decompositions. Following this approach, background modeling can be formulated as a matrix separation problem into low-rank (background) and sparse (foreground) components. This viewpoint has been advocated, for instance, by Cand\`{e}s et al.~\cite{RPCA1} in the framework of robust principal component analysis (RPCA). 
For a thorough discussion of such methods used for background modeling, we refer to Bouwmans et al. \cite{bouwmans2014robust,bouwmans2015decomp}. The connection between DMD and RPCA was first established by Grosek and Kutz \cite{grosek2014}. Assume the set of background modes $\{\omega_{p}\}$ satisfies $|\omega_{p}| \approx 0$. The DMD expansion of equation \eqref{eq:omegaj} then yields  
\begin{equation}
\label{equ:DMDTerms}
\begin{array}{ccccc}
{\bf X}_{\text{DMD}} & = & {\bf L} & $+$ & {\bf S} \\
& = & \underbrace{\sum_{p} b_{p}{\bf \phi}_{p} \lambda_p^{\mathbf{t}-1} }_{\text{Background Video}} & $+$  & \underbrace{\sum_{j \neq p} b_{j}{\bf \phi}_{j} \lambda_j^{\mathbf{t}-1} }_{\text{Foreground Video}} \\ 
\end{array} 
\end{equation}
where $\mathbf{t}=[1,...,m]$ is a $1 \times m$ time vector and ${\bf X}_{\text{DMD}} \in \mathbb{C}^{n \times m}$.\footnote{Note that by construction ${\bf X}_{\text{DMD}}$ is complex, while pixel intensities of the original video stream are real-valued. Hence, only the the real part is considered in the following.}
Specifically, DMD provides a matrix decomposition of the form  ${\bf X}_{DMD} = {\bf L} + {\bf S}$, where the low-rank matrix ${\bf L}$ will render the video of just the background, and the sparse matrix ${\bf S}$ will render the complementary video of the moving foreground objects. We can interpret these DMD results as follows: stationary background objects translate into highly correlated pixel regions from one frame to the next, which suggests a low-rank structure within the video data. Thus the DMD algorithm can be thought of as an RPCA method. The advantage of the DMD method and its sparse/low-rank separation is the computational efficiency of achieving (\ref{equ:DMDTerms}), especially when compared to the optimization methods of RPCA. 
The analysis of the time evolving amplitudes provide interesting opportunities. Specifically, learning the amplitudes' profiles for different foreground objects allows automatic separation of video feeds into different components. For instance, it could be of interest to discriminate between cars and pedestrians in a given video sequence.

\subsection{DMD for Real-Time Background Modeling}\label{sec:backModeling}
When dealing with high-resolution videos, the standard DMD approach is expensive in terms of computational time and memory, because the whole video sequence is reconstructed. Instead a `good' static background model is often sufficient for background subtraction. This is because background dynamics can be filtered out or thresholded. 
The challenge remains to automatically select the modes best describing the background. This is essentially a bias-variance trade-off. Using just the zero mode (background) leads to an under-fit background model, while a large set of modes tend to overfit. Motivated, by the sparsity-promoting variant of the standard DMD algorithm introduced by Jovanovi{\'c} et al. \cite{jovanovic2014sparsity}, we formulate a sparsity-constrained sparse coding problem for mode selection. The idea is to augment equation~\eqref{eq:b} by an additional term that penalizes the number of non-zero elements in the vector ${\bf b}$
\begin{equation}\label{eq:sparsecoding}
\hat{\bbeta}=\argmin_{\bf \bbeta} \|{\bf x}_1 - {\bf \Phi}{\bf \bbeta}\|_{F}^{2} \text{ such that } \|{\bf \bbeta}\|_0 < K,
\end{equation}  
where $\bbeta$ is the sparse representation of ${\bf b}$, and $\|\cdot \|_0$ is the $\ell_0$ pseudo norm which counts the non-zero elements in $\bbeta$. Solving this sparsity problem exactly is NP-hard. 
However, the problem in Eq.~\ref{eq:sparsecoding} can be efficiently solved using
greedy approximation methods. Specifically, we utilize orthogonal matching pursuit (OMP)~\cite{mallat1993matching,tropp2007signal}. A highly computationally efficient algorithm is proposed by Rubinstein et al.~\cite{rubinstein2008efficient} as implemented in the scikit-learn software package \cite{scikit-learn}. The greedy OMP algorithm works iteratively, selecting at each step the mode with the highest correlation to the current residual. Once a mode is selected the initial condition ${\bf x}_1$ is orthogonally projected on the span of the previously selected set of modes. Then the residual is recomputed and the process is repeated until $K$ non-zero entries are obtained. If no priors are available, the optimal number of modes $K$ can be determined using cross-validation. Finally, the background model is computed as 
\begin{equation}
\hat{\bf x}_{BG}={\bf \Phi}\hat{\bbeta}.
\end{equation} 
%
\section{Compressed DMD (cDMD)}
\label{sec:cdmd}
\begin{figure}[htp]
	\begin{center}
		\begin{tikzpicture}[thick,scale=0.85, every node/.style={transform shape}]
		\node [blocky4,name=fdata] { $\mathbf{X},\mathbf{X}'$};       
		\node [blocky4,name=fdmd, right of=fdata,node distance=5cm]  {$\mathbf{\Phi}, \mathbf{\Lambda}$};
		\node [blocky4,name=cdata,below of=fdata,node distance=3.5cm] {$\mathbf{Y},\mathbf{Y}'$};
		\node [blocky4,name=cdmd,below of=fdmd,node distance=3.5cm] {$\mathbf{\Phi_Y}, \mathbf{\Lambda_Y}$};
		
		\path [draw, ->, line width=1.2pt] (fdata) -- node[name=toparrow,above of=fdata,node distance=.25cm]{\textbf{DMD}}(fdmd);
		\path [draw, ->, line width=1.2pt] (cdata) -- node[name=bottomarrow,above of=cdata,node distance=.25cm]{\textbf{cDMD}}(cdmd);
		\path [draw, ->, line width=1.2pt] (fdata) -- node[name=leftarrow, left of=fdata, node distance = .25cm]{$\mathbf{C}$}(cdata);
		\path [draw, <-, line width=1.2pt] (fdmd) -- node[name=rightarrow, right of=fdmd, node distance = .75cm]{Eq.~\eqref{Eq:cDMDModes}}(cdmd);
		\node [name=data,above of=fdata,node distance=1.1cm] {\textbf{Data}};
		\node [name=modes,above of=fdmd,node distance=1.1cm] {\textbf{Dynamic Modes}};
		\node [name=full,left of=fdata,node distance=1.6cm]{\begin{sideways}\textbf{Full}\end{sideways}};
		\node [name=full,left of=cdata,node distance=1.6cm]{\begin{sideways}\textbf{Compressed}\end{sideways}};
		\end{tikzpicture}
	\end{center}
	\caption{Schematic of the compressed dynamic mode decomposition architecture. The data (video stream) is first compressed via left multiplication by a measurement matrix $\mathbf{C}$. DMD is then performed on the compressed representation of the data.  Finally, the full DMD modes $\mathbf{\Phi}$ are reconstructed from the compressed modes $\mathbf{\Phi}_Y$ by the expression in Eq.~\eqref{Eq:cDMDModes}.}\label{Fig:CSDMD}
\end{figure}
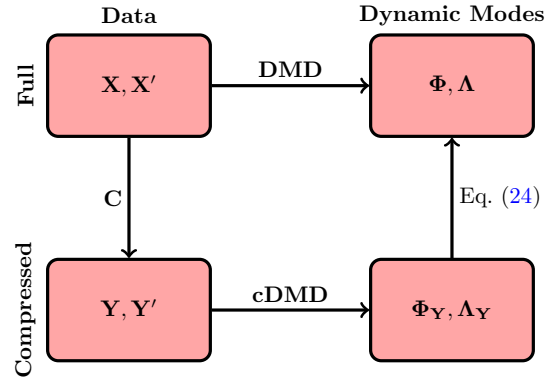

Compressed DMD provides a computationally efficient framework to compute the dynamic mode decomposition on massively under-sampled or compressed data~\cite{cdmd}.  
The method was originally devised to reconstruct high-dimensional, full-resolution DMD modes from sparse, spatially under-resolved measurements by leveraging compressed sensing.
However, it was quickly realized that if full-state measurements are available, many of the computationally expensive steps in DMD may be computed on a compressed representation of the data, providing dramatic computational savings.  
The first approach, where DMD is computed on sparse measurements without access to full data, is referred to as \emph{compressed sensing DMD}. 
The second approach, where DMD is accelerated using a combination of calculations on compressed data and full data, is referred to as \emph{compressed DMD} (cDMD); this is depicted schematically in Fig.~\ref{Fig:CSDMD}. 
For the applications explored in this work, we use compressed DMD, since full image data is available and reducing algorithm run-time is critical for real-time performance.  

\subsection{Compressed Sensing and Matrix Sketching}
\label{cs}
Compression algorithms are at the core of modern video, image and audio processing software such as MPEG, JPEG and MP3. In our mathematical infrastructure of compressed DMD, we consider the theory of compressed sensing and matrix sketching. 

\paragraph{Compressed sensing} demonstrates that instead of measuring the high-dimensional signal, or pixel space representation of a single frame ${\bf x}$, we can measure instead a low-dimensional subsample ${\bf y}$ and approximate/reconstruct the full state space ${\bf x}$ with this significantly smaller measurement~\cite{Donoho:2006,Candes2008,Baraniuk:2007}.  Specifically, compressed sensing assumes the data being measured is compressible in some basis, which is certainly the case for video.  Thus the video can be represented in a small number of elements of that basis, i.e. we only need to solve for the few non-zero coefficients in the transform basis.  For instance, 
consider the measurements ${\bf y}\in\mathbb{R}^p$, with $k< p\ll n$:
\begin{equation}\label{eq:yCx}
{\bf y} = {\bf C}{\bf x}.
\end{equation}
If ${\bf x}$ is sparse in ${\bf \Psi}$, then we may solve the underdetermined system of equations
\begin{equation}
{\bf y}={\bf C}{\bf \Psi}{\bf s} \label{eq:underdet}
\end{equation}
for ${\bf s}$ and then reconstruct ${\bf x}$.  Since there are infinitely many solutions to this system of equations, we seek the sparsest solution $\bhs$.  However, it is well known from the compressed sensing literature that solving for the sparsest solution formally involves an $\ell_0$ optimization that is NP-hard.  The success of compressed sensing is that it ultimately engineered a solution around this issue by showing that one can instead, under certain conditions on the measurement matrix ${\bf C}$, trade the
infeasible $\ell_0$ optimization for a convex $\ell_1$-minimization~\cite{Donoho:2006}:
\begin{equation}
\bhs=\argmin_{{\bf s}'}\|{\bf s}'\|_1,\text{ such that }{\bf y}={\bf C}{\bf \Psi}{\bf s}'.\label{eq:l1min} 
\end{equation}
Thus the $\ell_1$-norm acts as a proxy for sparsity promoting solutions of $\bhs$.  
To guarantee that the compressed sensing architecture will almost certainly work in a probabilistic sense, the measurement matrix ${\bf C}$ and sparse basis ${\bf \Psi}$ must be \emph{incoherent}, meaning that the rows of ${\bf C}$ are uncorrelated with the columns of ${\bf \Psi}$. This is discussed in more detail in~\cite{cdmd}. Given that we are considering video frames, it is easy to suggest the use of generic basis functions such as Fourier or wavelets in order to represent the sparse signal ${\bf s}$.  Indeed, wavelets are already the standard for image compression architectures such as JPEG-2000.  As for the Fourier transform basis, it is particularly attractive for many engineering purposes since single-pixel measurements are clearly incoherent given that it excites broadband frequency content. 

\paragraph{Matrix sketching} is another prominent framework in order to obtain a similar compressed representation of a massive data matrix~\cite{liberty2013simple,SketchingNLA}. The advantage of this approach are the less restrictive assumptions and the straight forward generalization from vectors to matrices. Hence, Eq.~\ref{eq:yCx} can be reformulated in matrix notation 
\begin{equation}\label{eq:YCX}
{\bf Y} = {\bf C}{\bf X},
\end{equation}
where again ${\bf C}$ denotes a suitable measurement matrix. Matrix sketching comes with interesting error bounds and is applicable whenever the data matrix ${\bf X}$ has low-rank structure. For instance, it has been successfully demonstrated that the singular values and right singular vectors can be approximated from such a compressed matrix representation~\cite{Gilbert2012}. 
\subsection{Algorithm}
\begin{figure}
	\centering
	\DeclareGraphicsExtensions{.pdf}
	\includegraphics[width=0.4\textwidth]{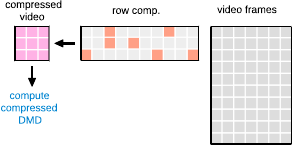}
	\caption{Video compression using a sparse measurement matrix. The compressed matrix faithfully captures the essential spectral information of the video.}
	\label{fig:dmdcompressed}
\end{figure}

The compressed DMD algorithm proceeds similarly to the standard DMD algorithm~\cite{tu2013dynamic} at nearly every step until the computation of the DMD modes. The key difference is that we first compute a compressed representation of the video sequence, as illustrated in Figure~\ref{fig:dmdcompressed}.
Hence the algorithm starts by generating the measurement matrix $\mathbf{C}\in\mathbb{R}^{p\times n}$ in order to compresses or sketch the data matrices as in Eq.~\eqref{Eq:FullData}:
\begin{equation}\label{(eq:YCX)}
\mathbf{Y} = \mathbf{C}\mathbf{X}, \quad\mathbf{Y}' = \mathbf{C}\mathbf{X}'.
\end{equation}
where $p$ is denoting the number of \emph{samples} or \emph{measurements}. There is a fundamental assumption that the input data are low-rank. This is satisfied for video data, because each of the columns of ${\bf X}$ and ${\bf X}' \in\mathbb{R}^{n\times m-1}$ are sparse in some transform basis ${\bf \Psi}$. Thus, for sufficiently many incoherent measurements, the compressed matrices $\bY$ and $\bY' \in\mathbb{R}^{p\times m-1}$ have similar correlation structures to their high-dimensional counterparts.
Then, compressed DMD approximates the eigenvalues and eigenvectors of the linear map $\mathbf{A_Y}$, where the estimator is defined as:
\begin{subequations}
	\begin{align}
	\mathbf{\hat{A}}_{\mathbf{Y}} &=\mathbf{Y}'\mathbf{Y}^\dagger\\
	&= \mathbf{Y}'\mathbf{V_Y}\boldsymbol{\bf S}_{\mathbf{Y}}^{-1}\mathbf{U_Y}^*,
	\end{align}
\end{subequations}
where $*$ denotes the conjugate transpose. The pseudo-inverse $\mathbf{Y}^\dagger$ is computed using the SVD:
\begin{equation}\label{eq:svd}
\mathbf{Y}=\mathbf{U_Y}\boldsymbol{\bf S}_{\mathbf{Y}}\mathbf{V_Y}^*,
\end{equation}
where the matrices ${\bf U}\in {\mathbb{R}}^{p\times k}$, and ${\bf V}\in {\mathbb{R}}^{m-1\times k}$ are the truncated left and right singular vectors. The diagonal matrix ${\bf S}\in {\mathbb{R}}^{k\times k}$ has the corresponding singular values as entries. Here $k$ is the target-rank of the truncated SVD approximation to ${\bf Y}$. 
Note that the subscript $\mathbf{Y}$ is included to explicitly denote computations involving the compressed data $\mathbf{Y}$.  
As in the standard DMD algorithm, we typically do not compute the large matrix $\mathbf{\hat{A}_Y}$, but instead compute the low-dimensional model projected onto the left singular vectors:
\begin{subequations}
	\begin{align}
	\tilde{\mathbf{A}}_{\mathbf{Y}} &= \mathbf{U_Y}^*\mathbf{\hat{A}_Y}\mathbf{U_Y}\\
	&= \mathbf{U_Y}^*\mathbf{Y}'\mathbf{V_Y}\boldsymbol{\bf S}_{\mathbf{Y}}^{-1}.
	\end{align}
\end{subequations}
Since this is a similarity transform, the eigenvectors and eigenvalues can be obtained from the eigendecomposition of $\tilde{\mathbf{A}}_{\mathbf{Y}}$
\begin{equation}
\tilde{\mathbf{A}}_{\mathbf{Y}}\mathbf{W_Y} = \mathbf{W_Y}\boldsymbol{\Lambda}_\mathbf{Y},
\end{equation}
where columns of ${\bf W_Y}$ are eigenvectors ${\bf \phi}_j$ and $\boldsymbol{\Lambda_Y}$ is a diagonal matrix containing the corresponding eigenvalues $\lambda_j$. The similarity transform implies that $\boldsymbol{\Lambda} \approx \boldsymbol{\Lambda_Y}$. The compressed DMD modes are consequently given by 
\begin{equation}
\mathbf{\Phi_Y} = \mathbf{Y}'\mathbf{V_Y}\boldsymbol{\bf S}_{\mathbf{Y}}^{-1}\mathbf{W_Y}.
\end{equation}
Finally, the full DMD modes are recovered using
\begin{equation}
\mathbf{\Phi} = \mathbf{X}'\mathbf{V_Y}\boldsymbol{\bf S}_{\mathbf{Y}}^{-1}\mathbf{W_Y}.\label{Eq:cDMDModes}
\end{equation}
%
%
Note that the compressed DMD modes in Eq.~\eqref{Eq:cDMDModes} make use of the full data $\mathbf{X}'$ as well as the linear transformations obtained using the compressed data $\mathbf{Y}$ and $\mathbf{Y}'$.  
The expensive SVD on $\mathbf{X}$ is bypassed, and it is instead performed on $\mathbf{Y}$.  
Depending on the compression ratio, this may provide significant computational savings.  
\begin{algorithm*}[htp]
	\begin{minipage}{210mm}
		\begin{tabbing}
			\hspace{10mm} \= \hspace{5mm} \= \hspace{5mm} \= \hspace{60mm} \=\kill
			\\
			\> \textbf{function} $[\mathbf{\Phi},\mathbf{b},\mathcal{V}] = \texttt{cdmd}(\mathbf{D}, k, p)$\\[3mm]
			(1)  \> \> $\mathbf{X,X^{\prime}} = \mathbf{D}$ \> \> {\color{blue}\textrm{Left/right snapshot sequence.}} \\[1mm]
			
			(2)  \> \> $\mathbf{C} = \texttt{rand}(p,m)$ \> \> {\color{blue}\textrm{Draw $p \times m$ sensing matrix.}}\\[1mm]
			
			(3)  \> \> $\mathbf{Y,Y^{\prime}}  = \mathbf{C} * \mathbf{D} $ \> \> {\color{blue}\textrm{Compress input matrix.}}\\[1mm]

			(4)  \> \> $\mathbf{U},\mathbf{S},\mathbf{V} = \texttt{svd}(\mathbf{Y}, k)$ \> \> {\color{blue}\textrm{Truncated SVD.}}
			\\[1mm]
			
			(6)  \> \> $\mathbf{\tilde{A}} = \mathbf{U}^{*} * \mathbf{Y^{\prime}} * \mathbf{V} * \mathbf{S}^{-1}$ \> \> {\color{blue}\textrm{Least squares fit.}} \\[1mm]
			
			(7)  \> \> $\mathbf{W},\mathbf{\Lambda} = \texttt{eig}(\mathbf{\tilde{A}})$  \> \> {\color{blue}\textrm{Eigenvalue decomposition.}} \\[1mm]
			
			(8)  \> \> $\mathbf{\Phi} \gets$ $\mathbf{X^{\prime}} * \mathbf{V} * \mathbf{S}^{-1} * \mathbf{W}$  \> \> {\color{blue}\textrm{Compute full-state modes $\mathbf{\Phi}$.}} \\[1mm]
			
			(9)  \> \> $\mathbf{b} = \texttt{lstsq}(\mathbf{\Phi},\mathbf{x}_{1})$  \> \> {\color{blue}\textrm{Compute amplitudes using $\mathbf{x}_{1}$ as intial condition.}} \\[1mm]
			
			(10)  \> \> $\mathcal{V} = \texttt{vander}(\texttt{diag}(\mathbf{\Lambda}))$  \> \> {\color{blue}\textrm{Vandermonde matrix (optional).}} \\[1mm]
			
		\end{tabbing}
	\end{minipage}
	\centering
	\caption{Compressed Dynamic Mode Decomposition. Given a matrix $\mathbf{D} \in \mathbb{R}^{n \times m}$ containing the flattened video frames, this procedure computes the approximate	dynamic mode decomposition, where $\mathbf{\Phi} \in \mathbb{C}^{n \times k}$ are the DMD modes,  $\mathbf{b} \in \mathbb{C}^{k}$ are the amplitudes, and $\mathcal{V} \in \mathbb{C}^{k \times m}$ is the Vandermonde matrix describing the temporal evolution. The procedure can be controlled by the two parameters $k$ and $p$, the target rank and the number of samples respectively. It is required that $n \geq m$, integer $k,p \geq 1$ and $k \ll n$ and $p \geq k$.}
	\label{Alg:cDMDalgorithm}			
\end{algorithm*}
The computational steps are summarized in Algorithm~\ref{Alg:cDMDalgorithm} and further numerical details are presented in~\cite{cdmd}.
\begin{remark} The computational performance heavily depends on the measurement matrix used to construct the compressed matrix, as described in the next section. For a practical implementation sparse or single pixel measurements (random row sampling) are favored. The latter most memory efficient methods avoids the generation of a large number of random numbers and the expensive matrix-matrix multiplication in step 3. 
\end{remark}
\begin{remark} One alternative to the predefined target-rank $k$ is the recent hard-thresholding algorithm of Gavish and Donoho~\cite{gavish}. This method can can be combined with step 4 to automatically determine the optimal target-rank.
\end{remark}
\begin{remark} As described in Section~\ref{sec:backModeling} step 9 can be replaced by the orthogonal matching pursuit algorithm, in order to obtain a sparsity-constrained solution: $\mathbf{b} = \texttt{omp}(\mathbf{\Phi},\mathbf{x}_{1})$. Computing the OMP solution is in general extremely fast, but if it comes to high resolution video streams this step can become computationally expensive. However, instead of computing the amplitudes based on the the full-state dynamic modes ${\bf \Phi}$ the compressed DMD modes ${\bf \Phi_Y}$ can be used. Hence, Eq.~\ref{eq:sparsecoding} can be reformulated as
	\begin{equation}
	\hat{\bbeta}=\argmin_{\bf \bbeta} \|{\bf y}_1 - {\bf \Phi_Y}{\bf \bbeta}\|_{F}^{2} \text{ such that } \|{\bf \bbeta}\|_0 < K,
	\end{equation}  
	where ${\bf y}_1$ is the first compressed video frame. Then step 9 can be replaced by: $\mathbf{beta} = \texttt{omp}(\mathbf{\Phi_Y},\mathbf{y}_{1})$. 
\end{remark}

\subsection{Measurement Matrices}
A basic sensing matrix $\mathbf{C}$ can be constructed by drawing $p\times n$ independent random samples from a Gaussian, Uniform or a sub Gaussian, e.g., Bernoulli distribution. It can be shown that these measurement matrices have optimal theoretical properties, however for practical large-scale applications they are often not feasible. This is because generating a large number of random numbers can be expensive and computing \eqref{(eq:YCX)} using unstructured dense matrices has a time complexity of $O(pnm)$. From a computational perspective it is favorable to build a structured random sensing matrix which is memory efficient, and enables the execution of fast matrix-matrix multiplications. For instance, Woolfe et al. \cite{woolfe2008fast} showed that the costs can be reduced to $O(log(p)nm)$ using a subsampled random Fourier transform (SRFT) sensing matrix
\begin{equation}
\mathbf{C} = \mathbf{R} \mathbf{F} \mathbf{D},
\end{equation}  
where $\mathbf{R}\in\mathbb{C}^{p\times n}$ draws $p$ random rows (without replacement) from the identity matrix $\mathbf{I}\in\mathbb{C}^{n\times n}$. $\mathbf{F}\in\mathbb{C}^{n\times n}$ is the unnormalized discrete Fourier transform with the following entries $\mathbf{F}(j,k)=exp(-2\pi i(j-1)(k-1)/m)$ and $\mathbf{D}\in\mathbb{C}^{n\times n}$ is a diagonal matrix with independent random diagonal elements uniformly distributed on the complex unit circle. 
While the SRFT sensing matrix has nice theoretical properties, the improvement from $O(pnm)$ to $O(log(p)nm)$ is not necessarily significant. In practice it is often sufficient to construct even simpler sensing matrices. An interesting approach making the matrix-matrix multiplication \eqref{(eq:YCX)} redundant is to use single-pixel measurements (random row-sampling)
\begin{equation}
\mathbf{C} = \mathbf{R}.
\end{equation}  
In a practical implementation this allows construction of the compressed matrix $\mathbf{Y}$ from choosing $p$ random rows without replacement from $\mathbf{X}$. Hence, only $p$ random numbers need to be generated and no memory is required for storing a sensing matrix $\mathbf{C}$.
A different approach is the method of sparse random projections \cite{achlioptas2003database}. The idea is to construct a sensing matrix $\mathbf{C}$ with identical independent distributed entries as follows
\begin{equation}
c_{ij} = \left \{
\begin{array}{c lc} 	
$1$ & \quad  \textrm{with prob. $\frac{1}{2s}$ } \\
$0$ & \quad  \textrm{with prob. $1-\frac{1}{s}$ },	\\
$-1$ & \quad  \textrm{with prob. $\frac{1}{2s}$ }
\end{array}
\right.
\end{equation} 
where the parameter $s$ controls the sparsity. While Achlioptas \cite{achlioptas2003database} has proposed the values $s=1,2$, Li et al. \cite{li2006very} showed that also very sparse (aggressive) sampling rates like $s={n}/{log(n)}$ achieve accurate results. Modern sparse matrix packages allow rapid execution of \eqref{(eq:YCX)}.
%

\subsection{GPU Accelerated Implementation}
\begin{figure}
	\centering
	\begin{subfigure}[t]{0.23\textwidth}
		\centering
		\DeclareGraphicsExtensions{.pdf}
		\includegraphics[height=1.9in]{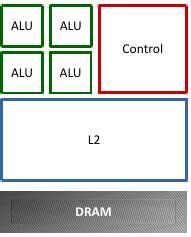}
		\caption{CPU}
	\end{subfigure}
	~~
	\begin{subfigure}[t]{0.23\textwidth}
		\centering
		\DeclareGraphicsExtensions{.pdf}
		\includegraphics[height=1.9in]{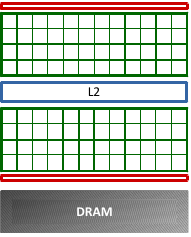}
		\caption{GPU}
	\end{subfigure}
	\caption{Illustration of the CPU and GPU architecture.}
	\label{fig:CPUGPU}
	
	\vspace{+0.2in}	
	\centering
	\DeclareGraphicsExtensions{.pdf}
	\includegraphics[width=0.3\textwidth]{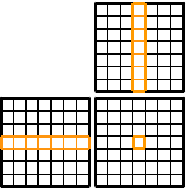}
	\caption{Illustration of the data parallelism in matrix-matrix multiplications.}
	\label{fig:GPUmat}	
\end{figure}

While most current desktop computers allow multithreading and also multiprocessing, using a graphics processing unit (GPU) enables massive parallel processing. The paradigm of parallel computing becomes more important as larger amounts of data stagnate CPU clock speeds. The architecture of a modern CPU and GPU is illustrated in Figure~\ref{fig:CPUGPU}. The key difference between these architectures is that the CPU consists of few arithmetic logic units (ALU) and is highly optimized for low-latency access to cached data sets, while the GPU is optimized for data-parallel, throughput computations. This is achieved by the large number of small arithmetic logic units (ALU).
Traditionally this architecture was designed for the real-time creation of high-definition 2D/3D graphics. However, NVIDIA's programming model for parallel computing \textit{CUDA} opens up the GPU as a general parallel computing device \cite{CUDA}. Using high-performance linear algebra libraries, e.g. \textit{CULA} \cite{CULA}, can help to accelerate comparable CPU implementations substantially. Take for instance the matrix multiplication of two $n\times n$ square matrices, illustrated in Figure~\ref{fig:GPUmat}. The computation involves the evaluation of $n^{2}$ dot products.\footnote{Modern efficient matrix-matrix multiplications are based on block matrix decomposition or other computational tricks, and do not actually compute $n^{2}$ dot products. However the concept of parallelism remains the same.} The data parallelism therein is that each dot-product can be computed independently. With enough ALUs the computational time can be substantially accelerated. This parallelism applies readily to the generation of random numbers and many other linear algebra routines.

Relatively few GPU accelerated background subtraction methods have been proposed \cite{carr2008gpu,pham2010gpu,Qin20151}. The authors achieve considerable speedups compared to the corresponding CPU implementations. However, the proposed methods barely exceed $25$ frames per second for high definition videos. This is mainly due to the fact that many statistical methods do not fully benefit from the GPU architecture. In contrast, linear algebra based methods can substantially benefit from parallel computing. An analysis of Algorithm~\ref{Alg:cDMDalgorithm} reveals that generating random numbers in line 2 and the dot products in lines 3, 6, and 8 are particularly suitable for parallel processing. But also the computation of the deterministic SVD, the eigenvalue decomposition and the least-square solver can benefit from the GPU architecture. Overall the GPU accelerated DMD implementation is substantially faster than the \textit{MKL} (Intel Math Kernel Library) accelerated routine. The disadvantage of current GPUs is the rather limited bandwidth, i.e., the amount of data which can be exchanged per unit of time, between CPU and GPU memory. However, this overhead can be mitigated using asynchronous memory operations.

\section{Results}
\label{sec:results}
In this section we evaluate the computational performance and the suitability of compressed DMD for object detection. 
To evaluate the detection performance, a foreground mask $\mathcal{X}$ is computed by thresholding the difference between the true frame and the reconstructed background. A standard method is to use the Euclidean distance, leading to the following binary classification problem
\begin{equation}\label{eq:thres}
\mathcal{X}_{t}(j) = \left \{
\begin{array}{lc lc} 	
1 && \quad  \textrm{if $\|x_{jt}-\hat{x}_{j}\| > \tau$}, \\
0 && \quad  \textrm{otherwise}	
\end{array}
\right.
\end{equation}
where $x_{jt}$ denotes the $j$-th pixel of the $t$-th video frame and $\hat{x}_{j}$ denotes the corresponding pixel of the modeled background. Pixels belonging to foreground objects are set to 1 and 0 otherwise. Access to the true foreground mask allows the computation of several statistical measures. For instance, common evaluation measures in the background subtraction literature are recall, precision and the F-measure. While recall measures the ability to correctly detect pixels belonging to moving objects, precision measures how many predicted foreground pixels are actually correct, i.e., false alarm rate. The F-measure combines both measures by their harmonic mean.
A workstation (Intel Xeon CPU E5-2620 2.4GHz, 32GB DDR3 memory and NVIDIA GeForce GTX 970) was used for all following computations.

\subsection{Evaluation on Real Videos}
\begin{figure}[h]
	\centering
	\captionsetup[subfigure]{labelformat=empty}
	\begin{subfigure}[t]{0.1\textwidth}
		\centering
		\DeclareGraphicsExtensions{.pdf}
		\includegraphics[height=0.7in]{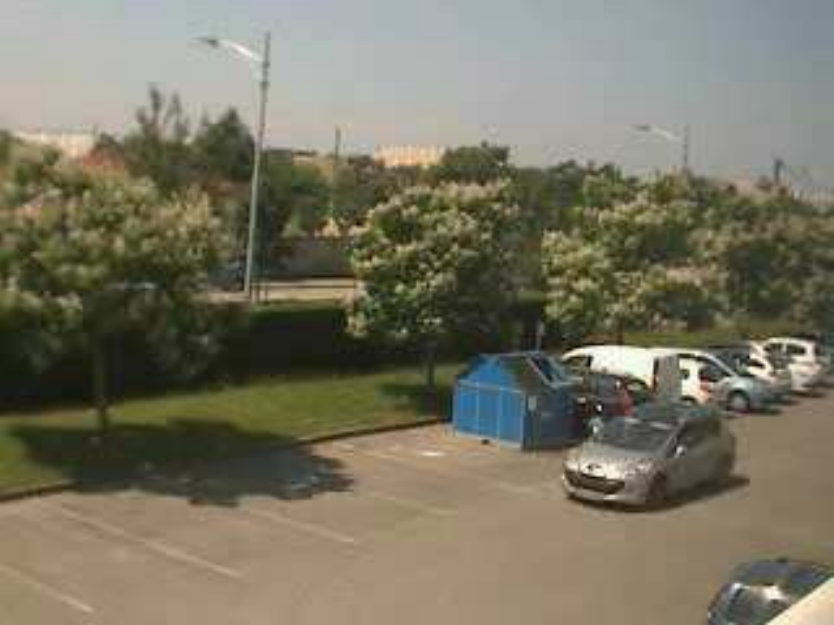}
		\caption{(001) Boring parking}
	\end{subfigure}
	~~~~~~
	\begin{subfigure}[t]{0.1\textwidth}
		\centering
		\DeclareGraphicsExtensions{.pdf}
		\includegraphics[height=0.7in]{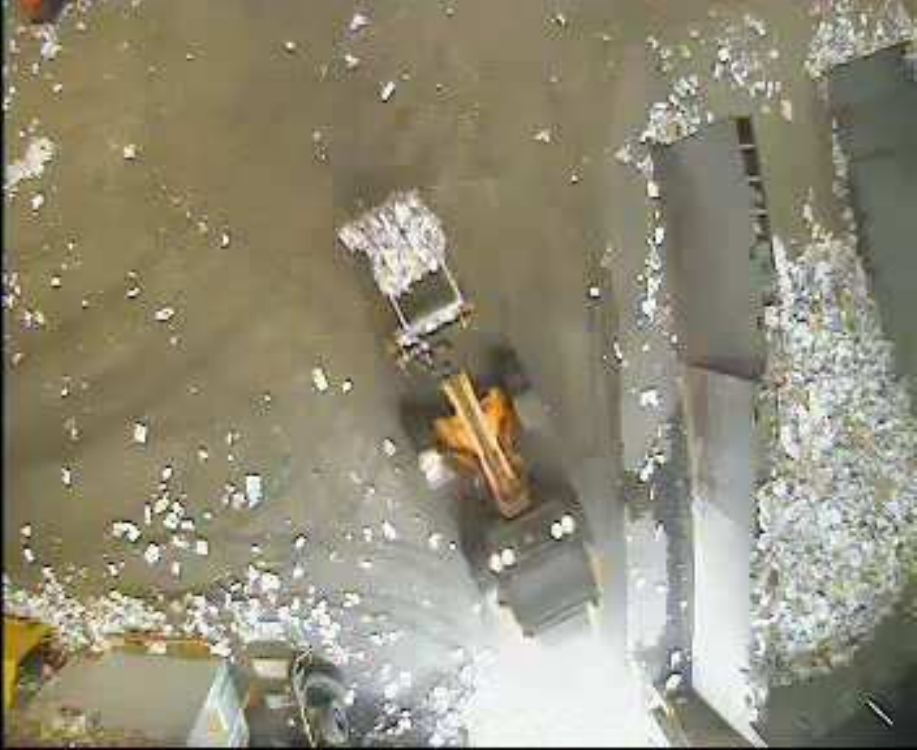}
		\caption{(002) Big trucks}
	\end{subfigure}
	~~~~~~
	\begin{subfigure}[t]{0.1\textwidth}
		\centering
		\DeclareGraphicsExtensions{.pdf}
		\includegraphics[height=0.7in]{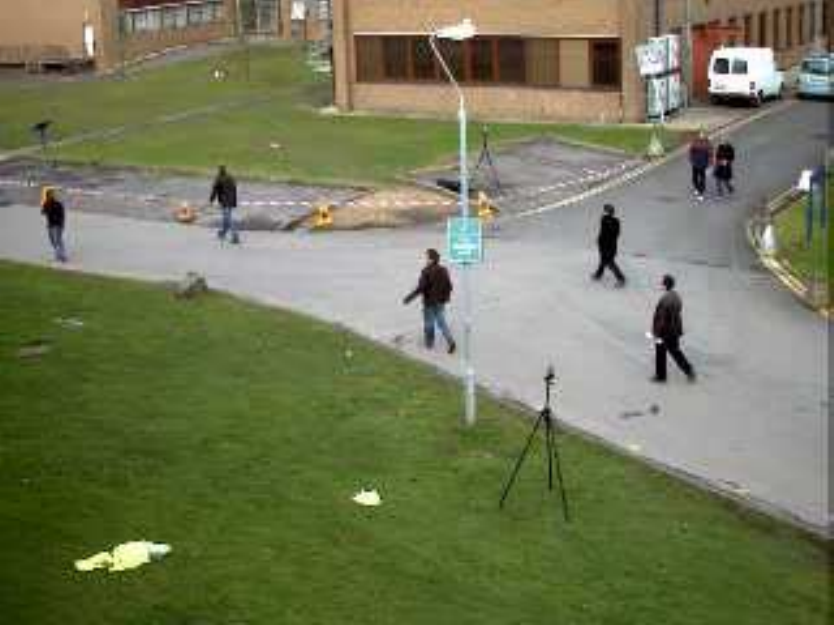}
		\caption{(003) Wandering students}
	\end{subfigure}	
	
	\begin{subfigure}[t]{0.1\textwidth}
		\centering
		\DeclareGraphicsExtensions{.pdf}
		\includegraphics[height=0.7in]{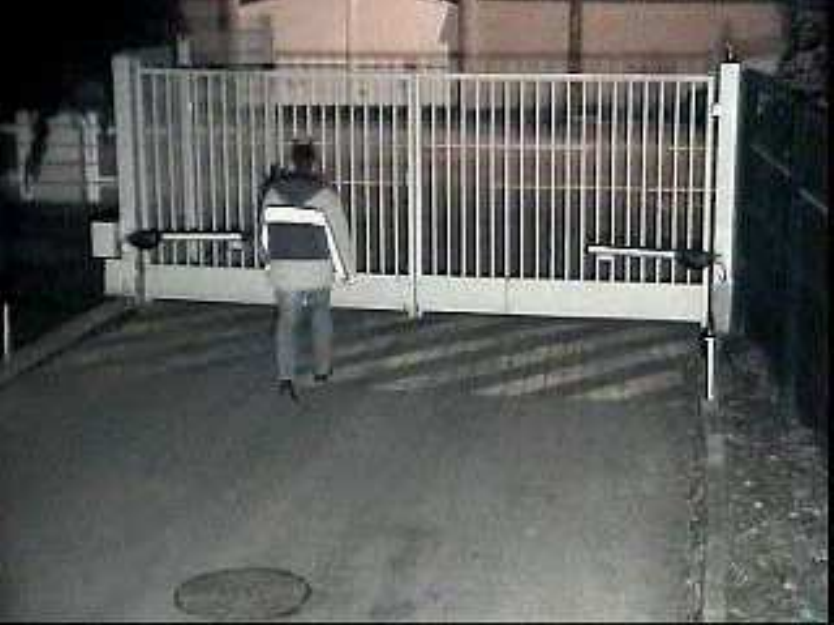}
		\caption{(004) Rabbit in the night}
	\end{subfigure}
	~~~~~~
	\begin{subfigure}[t]{0.1\textwidth}
		\centering
		\DeclareGraphicsExtensions{.pdf}
		\includegraphics[height=0.7in]{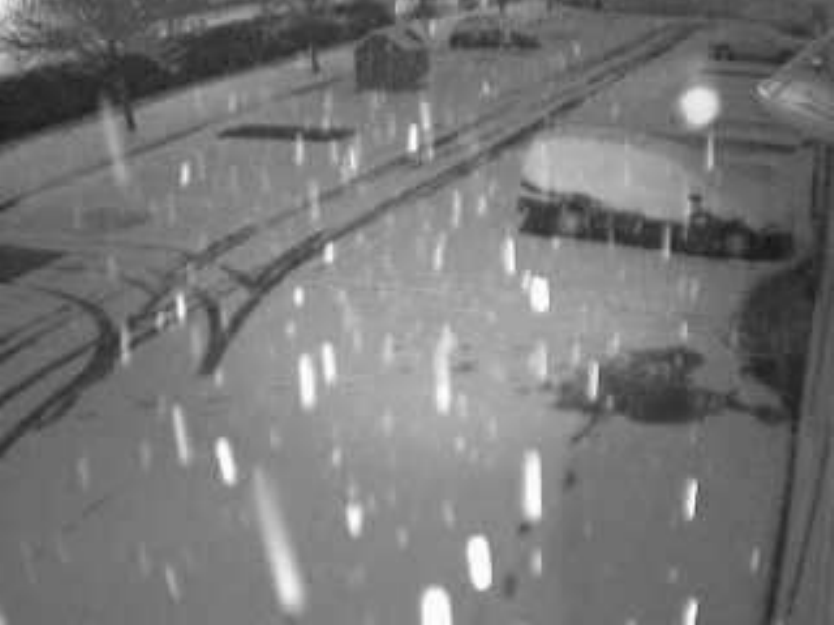}
		\caption{(005) Snowy Christmas}
	\end{subfigure}
	~~~~~~
	\begin{subfigure}[t]{0.1\textwidth}
		\centering
		\DeclareGraphicsExtensions{.pdf}
		\includegraphics[height=0.7in]{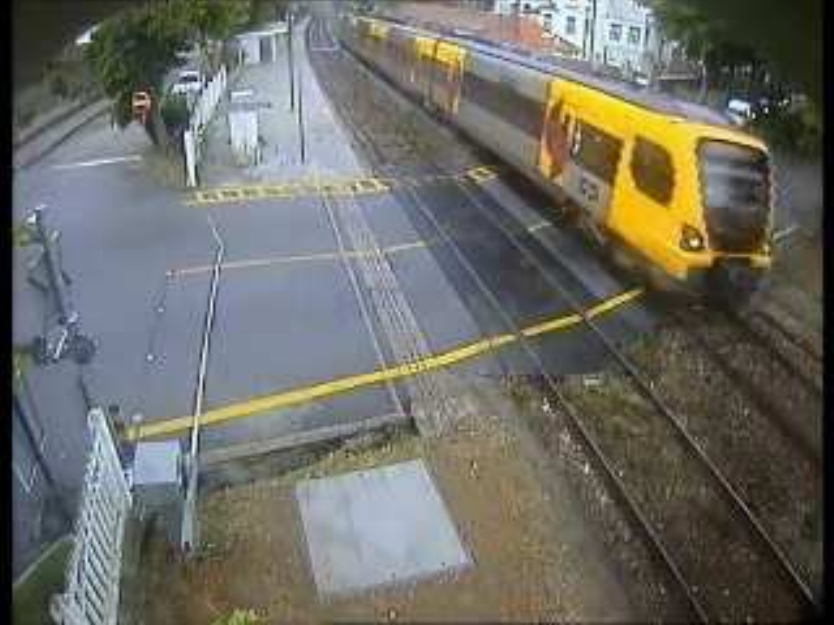}
		\caption{(006) Beware of the trains}
	\end{subfigure}
	
	\begin{subfigure}[t]{0.1\textwidth}
		\centering
		\DeclareGraphicsExtensions{.pdf}
		\includegraphics[height=0.7in]{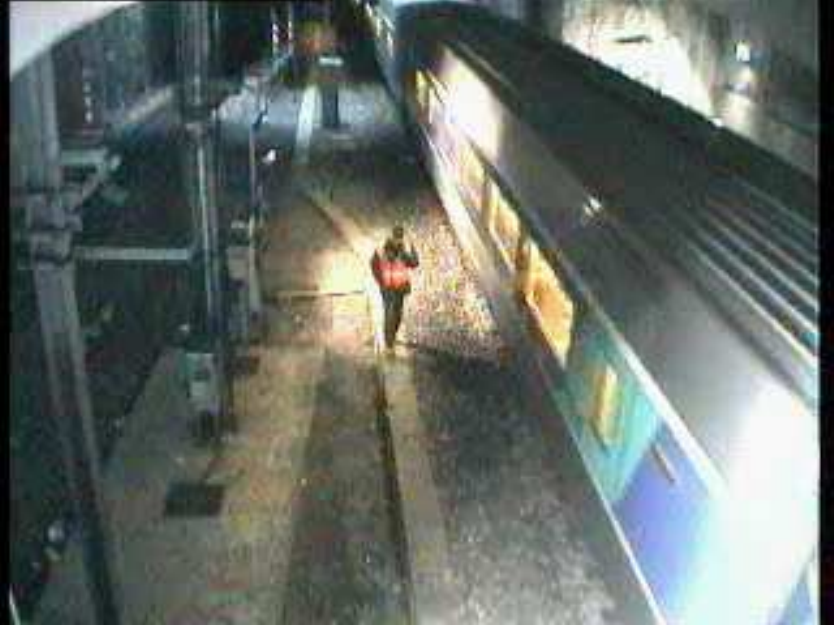}
		\caption{(007) Train in the tunnel}
	\end{subfigure}
	~~~~~~
	\begin{subfigure}[t]{0.1\textwidth}
		\centering
		\DeclareGraphicsExtensions{.pdf}
		\includegraphics[height=0.7in]{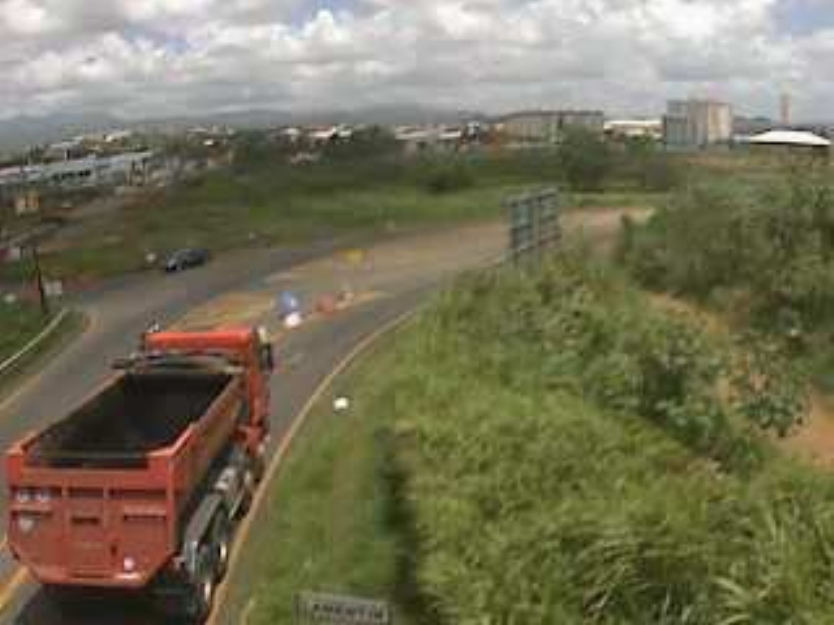}
		\caption{(008) Traffic during windy day’}
	\end{subfigure}
	~~~~~~
	\begin{subfigure}[t]{0.1\textwidth}
		\centering
		\DeclareGraphicsExtensions{.pdf}
		\includegraphics[height=0.7in]{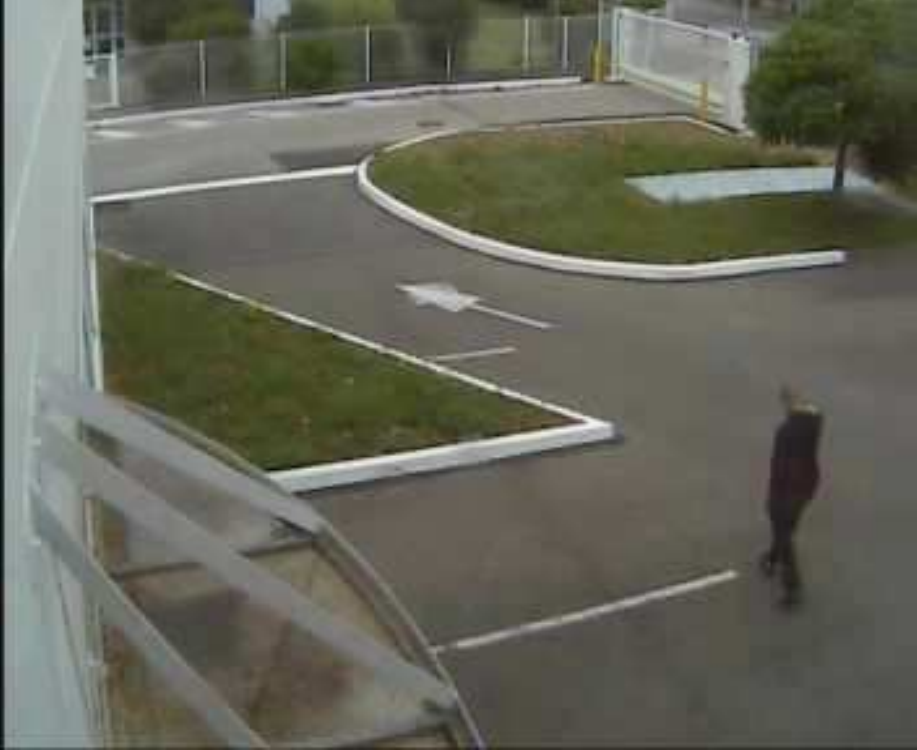}
		\caption{(009) One rainy hour}
	\end{subfigure}	
	\caption{BMC dataset: Example frames of the $9$ real videos.}
	\label{fig:exFrames}
\end{figure}

We have evaluated the performance of compressed DMD for object detection using the CD (ChangeDetection.net) and BMC (Background Models Challenge) benchmark dataset \cite{wang2014cdnet,vacavant2013benchmark}. Figure~\ref{fig:exFrames} illustrates the $9$ real videos of the latter dataset, posing many common challenges faced in outdoor video surveillance scenarios.
Mainly, the following complex situations are encountered:
\begin{itemize}
	\item \textbf{Illumination changes}: Gradual illumination changes caused by fog or sun.
	\item \textbf{Low illumination}: Bad light conditions, e.g., night videos.
	\item \textbf{Bad weather}: Introduced noise (small objects) by weather conditions, e.g., snow or rain.
	\item \textbf{Dynamic backgrounds}: Moving objects belonging to the background, e.g. waving trees or clouds. 
	\item \textbf{Sleeping foreground objects}: Former foreground objects that becoming motionless and moving again at a later point in time.
\end{itemize}
\begin{figure*}[htp]
	\centering
	\begin{subfigure}[t]{0.29\textwidth}
		\centering
		\DeclareGraphicsExtensions{.pdf}
		\includegraphics[width=1\textwidth]{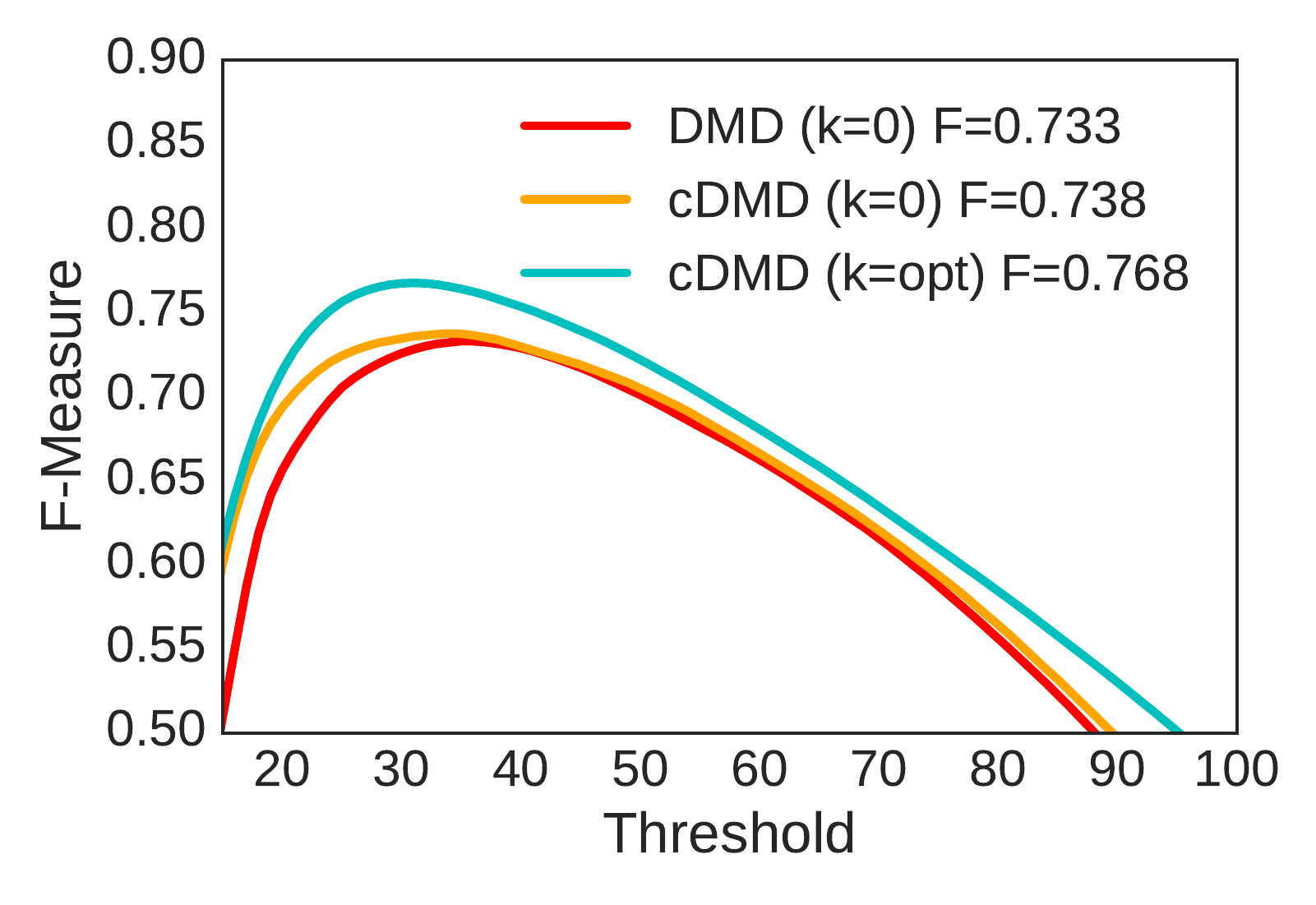}
		\caption{Highway}
	\end{subfigure}
	~~~~~~
	\begin{subfigure}[t]{0.29\textwidth}
		\centering
		\DeclareGraphicsExtensions{.pdf}
		\includegraphics[width=1\textwidth]{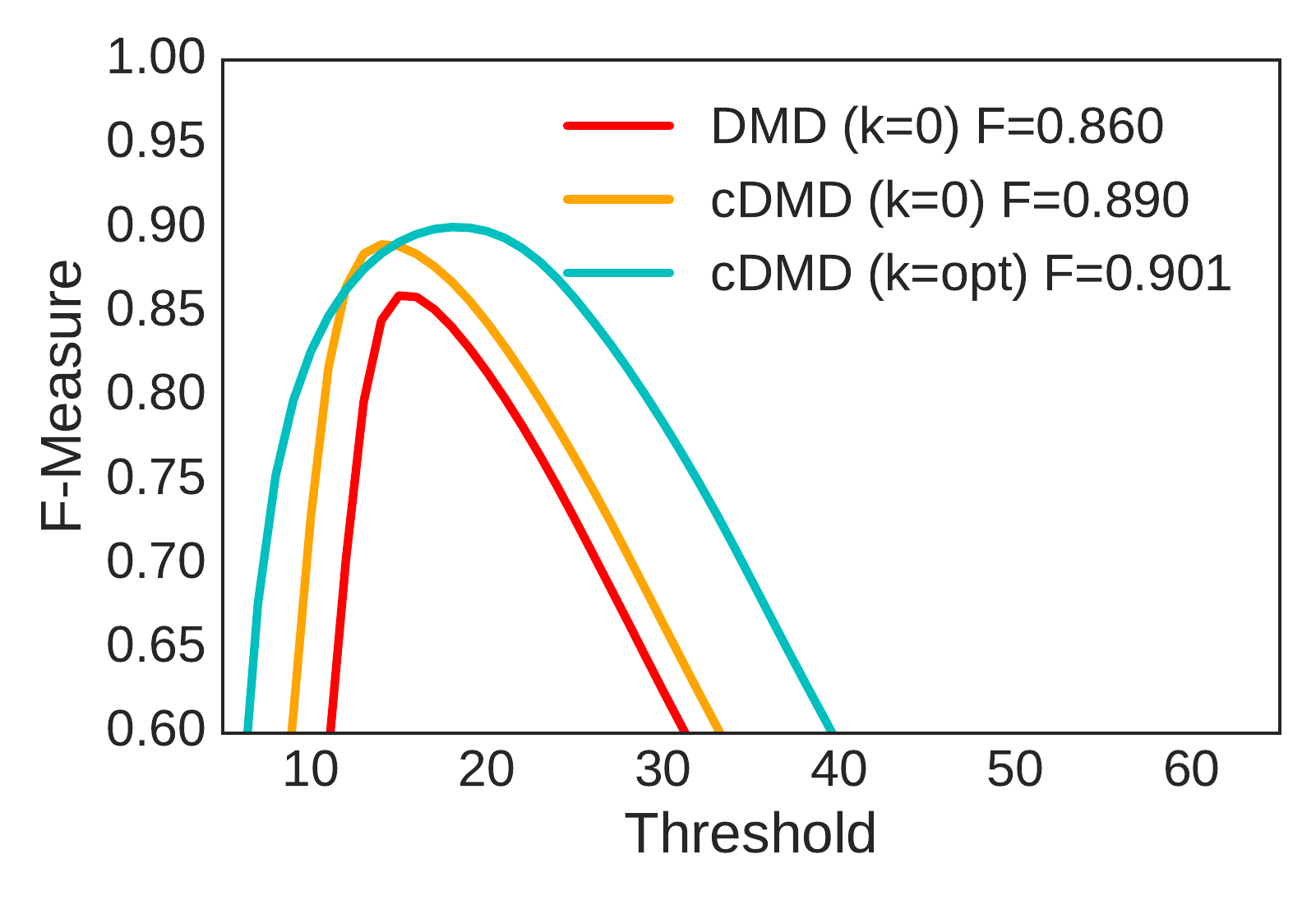}
		\caption{Blizzard}
	\end{subfigure}
	~~~~~~
	\begin{subfigure}[t]{0.29\textwidth}
		\centering
		\DeclareGraphicsExtensions{.pdf}
		\includegraphics[width=1\textwidth]{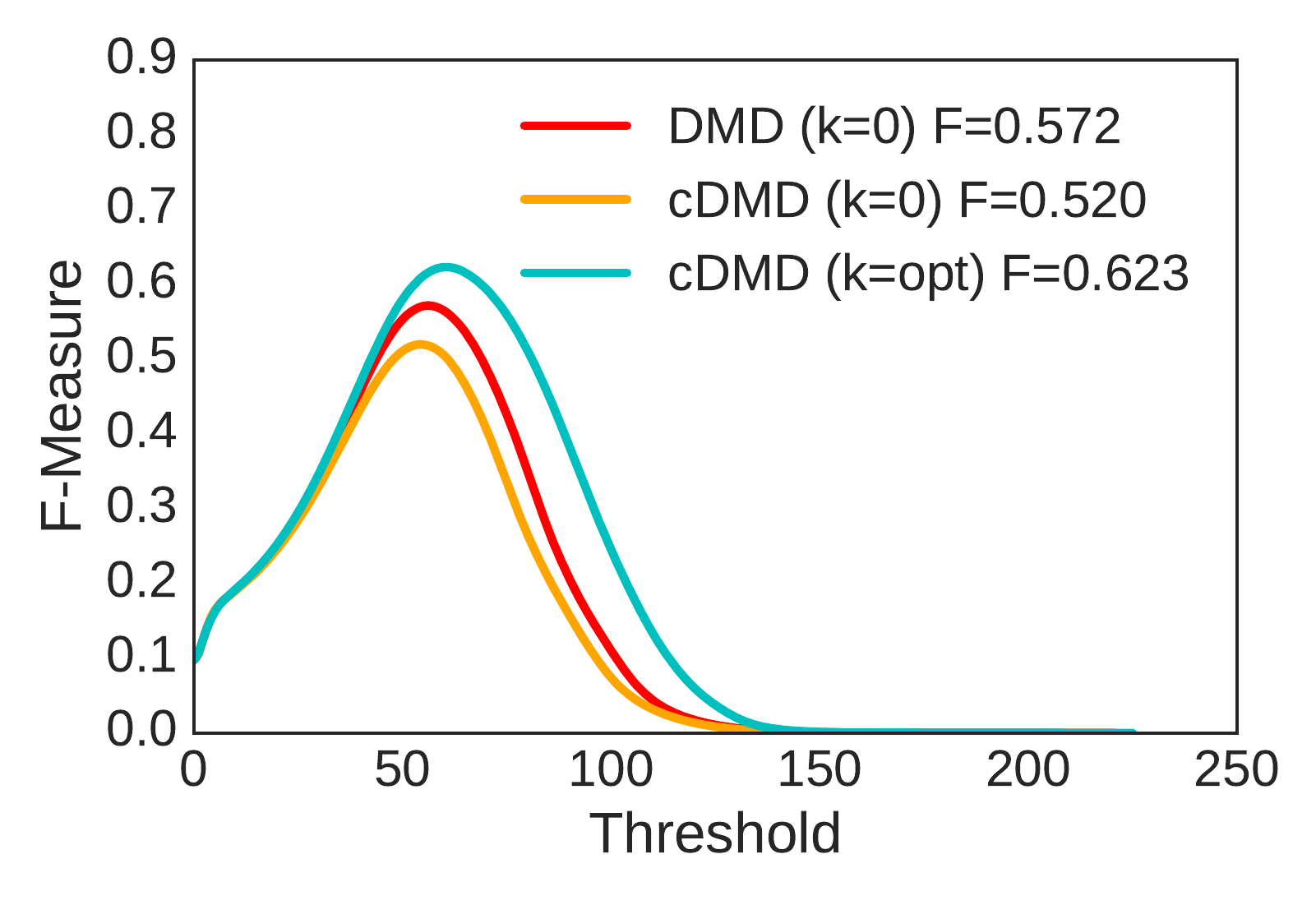}
		\caption{Canoe}
	\end{subfigure}	

	\begin{subfigure}[t]{0.29\textwidth}
		\centering
		\DeclareGraphicsExtensions{.pdf}
		\includegraphics[width=1\textwidth]{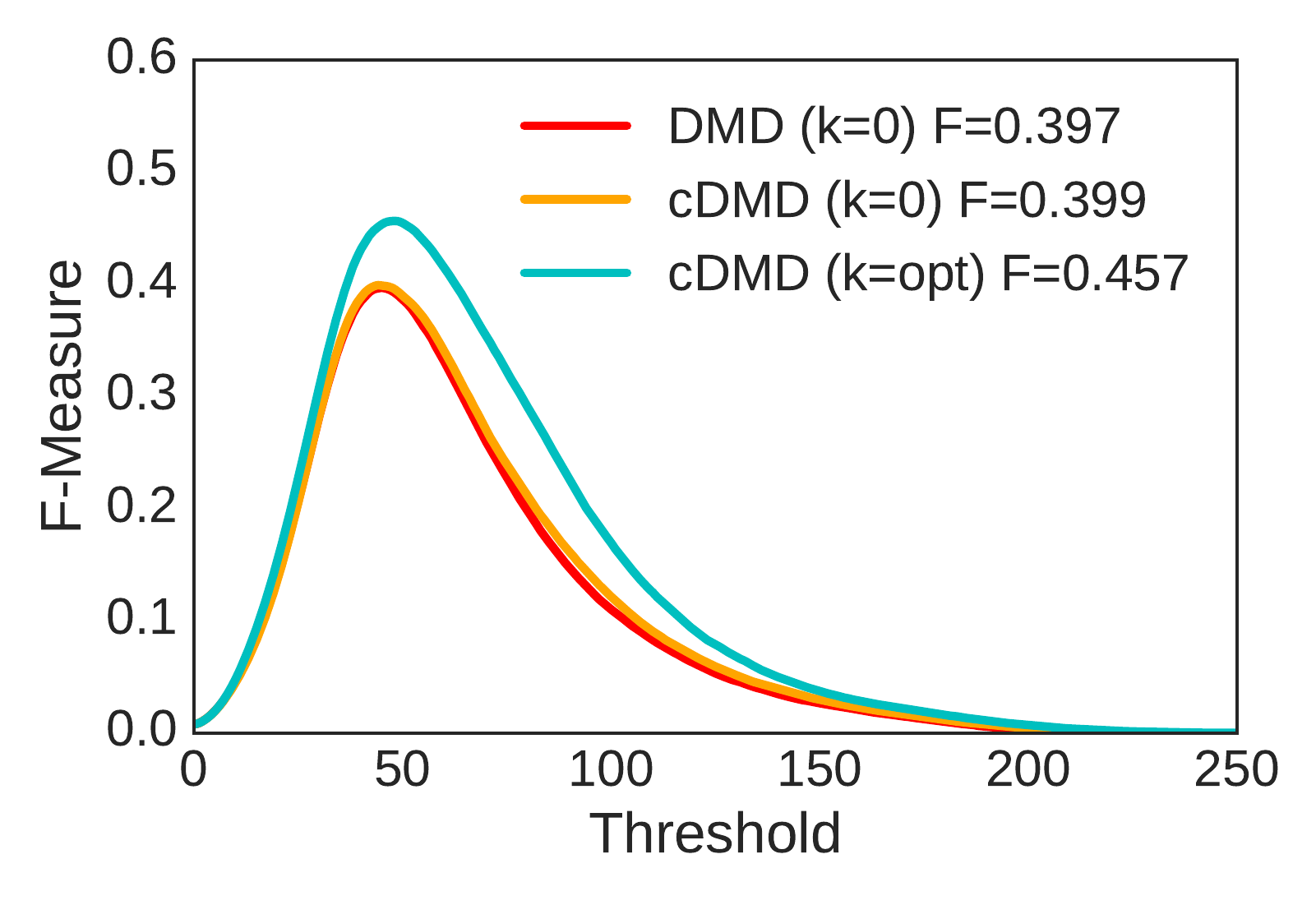}
		\caption{Fountain02}
	\end{subfigure}
	~~~~~~
	\begin{subfigure}[t]{0.29\textwidth}
		\centering
		\DeclareGraphicsExtensions{.pdf}
		\includegraphics[width=1\textwidth]{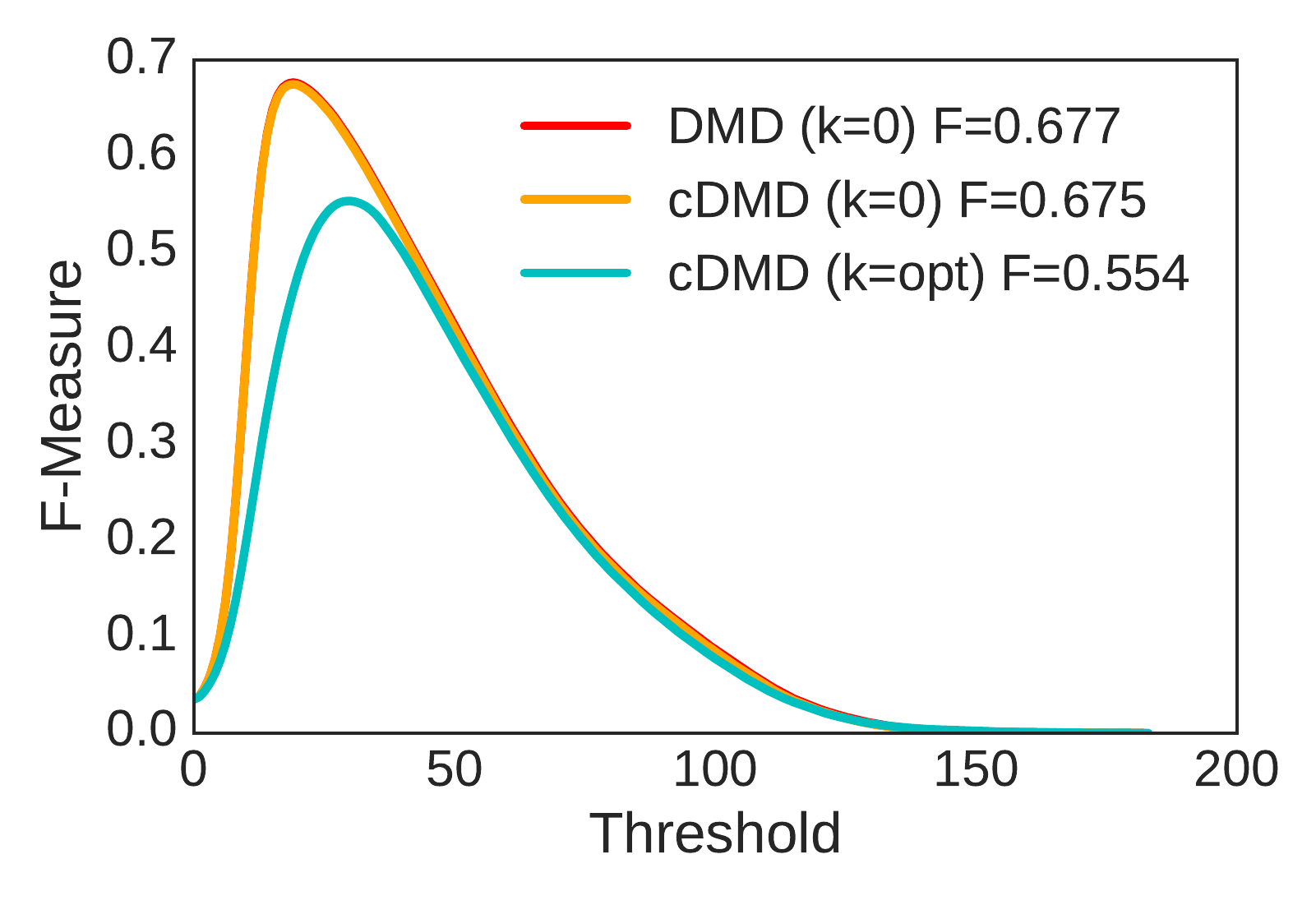}
		\caption{Park}
	\end{subfigure}
	~~~~~~
	\begin{subfigure}[t]{0.29\textwidth}
		\centering
		\DeclareGraphicsExtensions{.pdf}
		\includegraphics[width=1\textwidth]{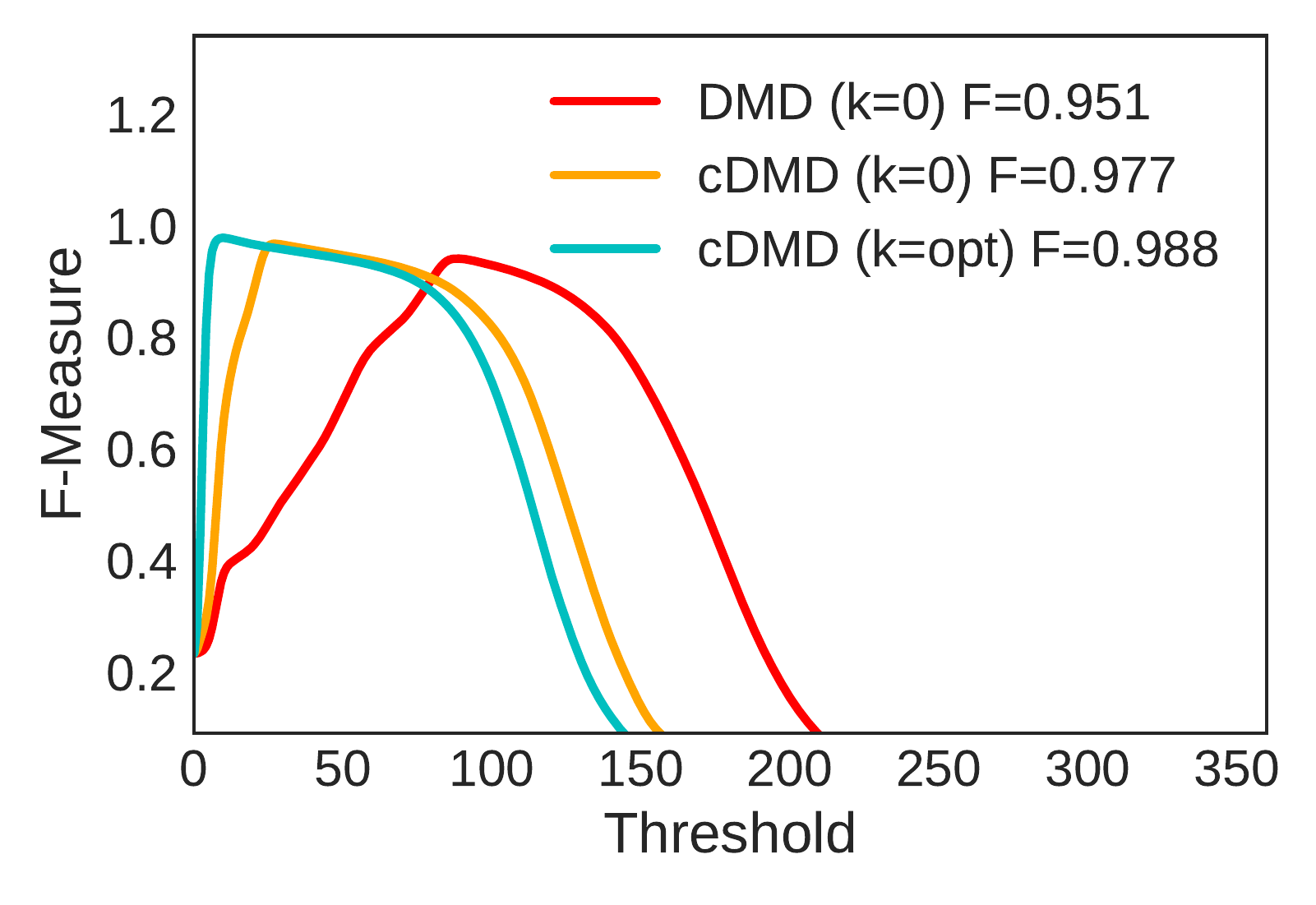}
		\caption{Library}
	\end{subfigure}
	\caption{The F-measure for varying thresholds is indicating the dominant background modeling performance of the sparsity-promoting compressed DMD algorithm. In particular, the performance gain (over using the zero mode only) is substantial for the dynamic background scenes `Canoe' and `Fountain02'.}
	\label{fig:CDnet}
\end{figure*}

\paragraph{Evaluation settings.} In order to obtain reproducible results the following settings have been used. For a given video sequence, the low-rank dynamic mode decomposition is computed using a very sparse measurement matrix with a sparsity factor $s=n/log(n)$ and $p=1000$ measurements. While, we use here a fixed number of samples, the choice can be guided by the formula $p > k\cdot \log(n/k)$. The target-rank $k$ is automatically determined via the optimal hard-threshold for singular values~\cite{gavish}. Once the dynamic mode decomposition is obtained, the optimal set of modes is selected using the orthogonal matching pursuit method. In general the use of $K=10$ non-zero entries achieves good results. Instead of using a predefined value for $K$, cross-validation can be used to determine the optimal number of non-zero entries. Further, the dynamic mode decomposition as presented here is formulated as a batch algorithm, in which a given long video sequence is split into batches of $200$ consecutive frames. The decomposition is then computed for each batch independently.   

\paragraph{The CD dataset.} First, six CD video sequences are used to contextualize the background modeling quality using the sparse-coding approach. This is compared to using the zero (static background) mode only. Figure~\ref{fig:CDnet} shows the evaluation results of one batch by plotting the F-measure against the threshold for background classification. In fife out of the six examples the sparse-coding approach (cDMD k=opt) dominates. In particular, significant improvements are achieved for the dynamic background video sequences `Canoe' and `Fountain02'. Only in case of the `Park' video sequence the method tends to over-fit. Interestingly, the performance of the compressed algorithm is slightly better then the exact DMD algorithm, overall. This is due to the implicit regularization of randomized algorithms~\cite{Mahoney2011,rSVDR}.

\begin{figure}[htp]
	\centering
	\DeclareGraphicsExtensions{.pdf}
	\includegraphics[width=0.5\textwidth]{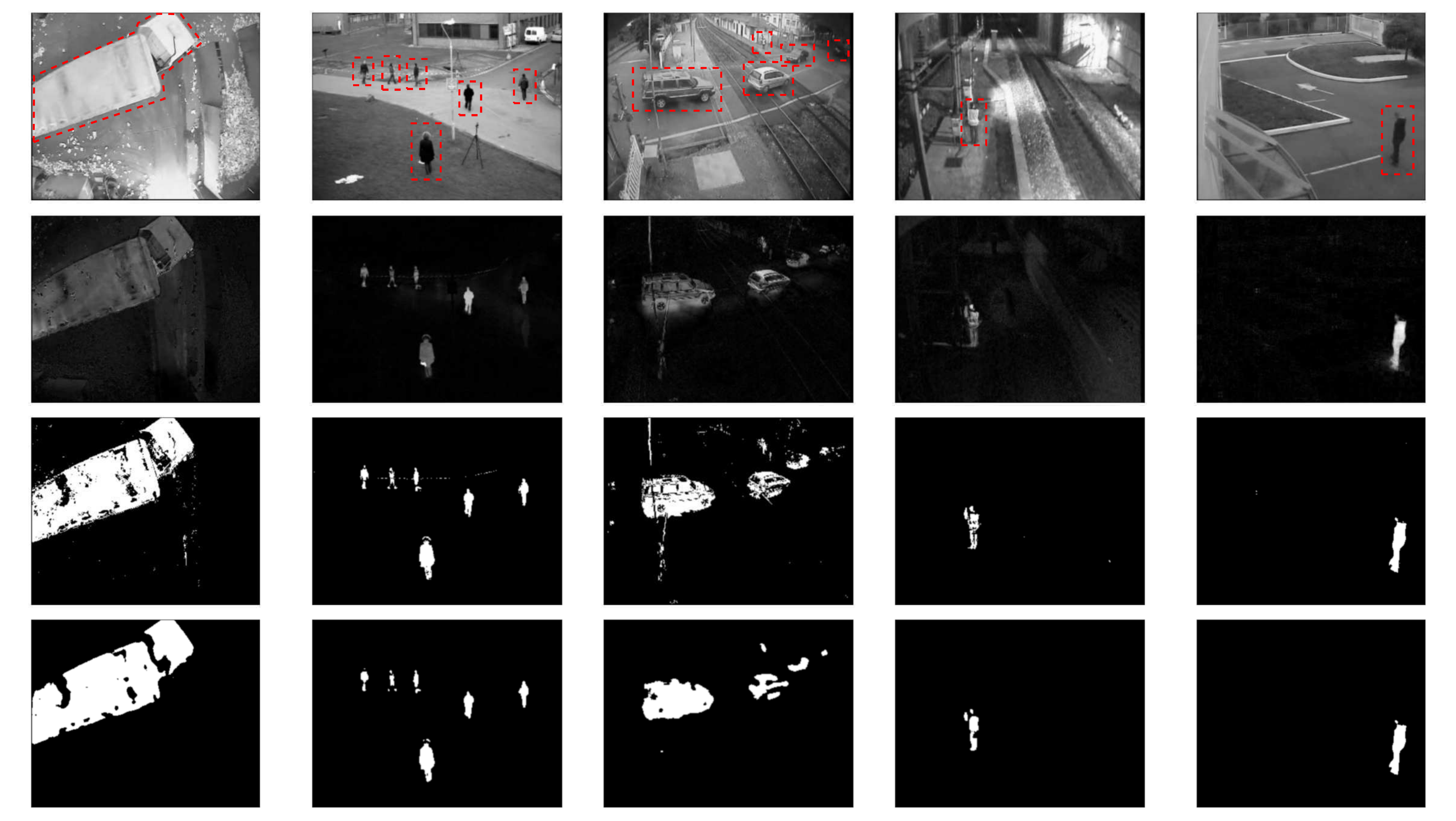}
	\caption{Visual evaluation results for 5 example frames corresponding to the BMC Videos: 002, 003, 006, 007 and 009. The top row shows the original grayscale images (moving objects are highlighted). The second row shows the differencing between the reconstructed cDMD background and the original frame. Row three shows the thresholded and row four the in addition median filtered foreground mask. }
	\label{fig:evalMat}
\end{figure}

\paragraph{The BMC dataset.} In order to compare the cDMD algorithm with other RPCA algorithms the BMC dataset has been used. Table~\ref{Tab:evaluationBMC} shows the evaluation results computed with the BMC wizard for all $9$ videos. An individual threshold value has been selected for each video to compute the foreground mask. For comparison the evaluation results of $3$ other RPCA methods are shown~\cite{bouwmans2015decomp}. Overall cDMD achieves an average F-value of about $0.648$. This is slightly better then the performance of GoDec \cite{Zhou11godec} and nearly as good as LSADM \cite{Goldfarb}. However, it is lower then the F-measure achieved with the RSL method \cite{RPCA2}. Figure~\ref{fig:evalMat} presents visual results for example frames across 5 videos. The last row shows the smoothed (median filtered) foreground mask.
\paragraph{Discussion.} The results reveal some of the strengths and limitations of the compressed DMD algorithm. First, because cDMD is presented here as a batch algorithm, detecting sleeping foreground objects as they occur in video 001 is difficult. Another weakness is the limited capability of dealing with non-periodic dynamic backgrounds, e.g., big waving trees and moving clouds as occurring in the videos 001, 005, 008 and 009. On the other hand good results are achieved for the videos 002, 003, 004 and 007, showing that DMD can deal with large moving objects and low illumination conditions. The integration of compressed DMD into a video system can overcome some of these initial issues. Hence, instead of discarding the previous modeled background frames, a background maintenance framework can be used to incrementally update the model. In particular, this allows to deal better with sleeping foreground objects. Further, simple post-processing techniques (e.g. median filter or morphology transformations) can substantially reduce the false positive rate. 

\subsection{Computational Performance}
\begin{figure*}[!htpb]
	\centering
	\DeclareGraphicsExtensions{.pdf}
	\includegraphics[width=0.95\textwidth]{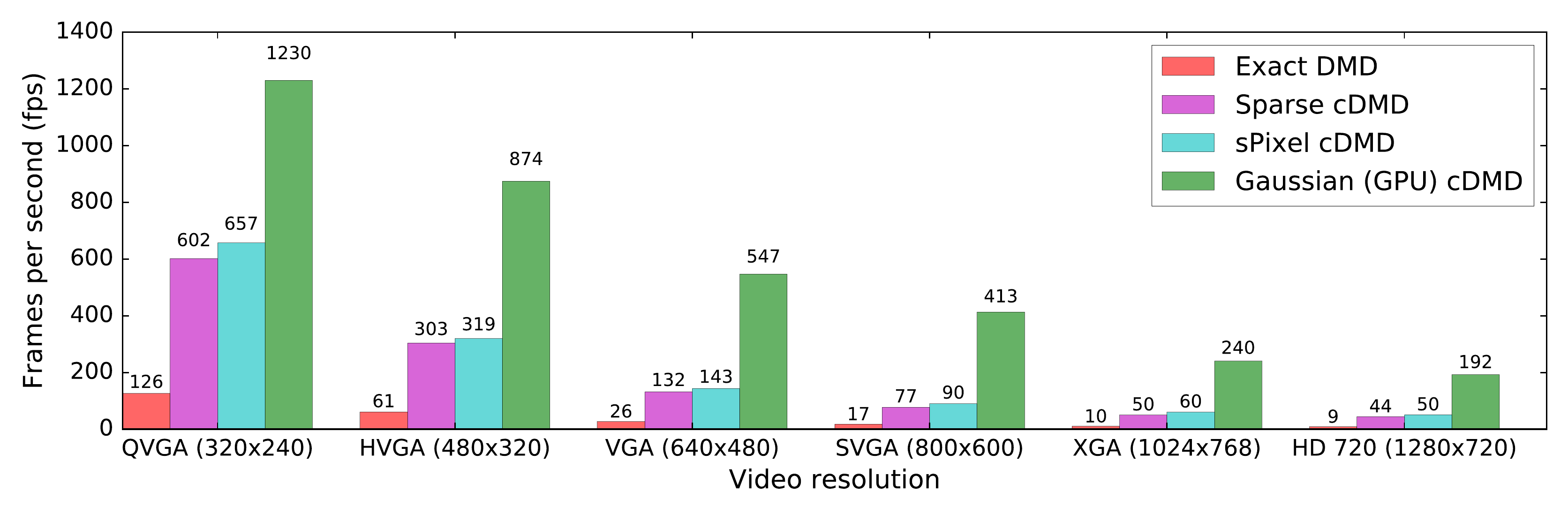}
	\caption{CPU and GPU algorithms runtime (including the computation of the foreground mask) for varying video resolutions (200 frames). The optimal target rank is automatically determined and $p=1000$ samples are used.}
	\label{fig:cDMDtiming}
\end{figure*}

Figure~\ref{fig:cDMDtiming} shows the average frames per seconds (fps) rate required to obtain the foreground mask for varying video resolutions. The results illustrate the substantial computational advantage of the cDMD algorithm over the standard DMD. The computational savings are mainly achieved by avoiding the expensive computation of the singular value decomposition. Specifically, the compression step reduces the time complexity from $O(knm)$ to $O(kpm)$. The computation of the full modes $\mathbf{\Phi}$ in Eq.~\ref{Eq:cDMDModes} remain the only computational expensive step of the algorithm. However, this step is embarrassingly parallel and the computational time can be further reduced using a GPU accelerated implementation. The decomposition of a HD $1280\times 720$ videos feed using the GPU accelerated implementation achieves a speedup of about $4$ and $21$ compared to the corresponding CPU cDMD and (exact) DMD implementations. The speedup of the GPU implementation can even further be increased using sparse or single pixel (sPixel) measurement matrices. 

\begin{figure}[htp]
	\centering
	\DeclareGraphicsExtensions{.pdf}
	\includegraphics[width=0.5\textwidth]{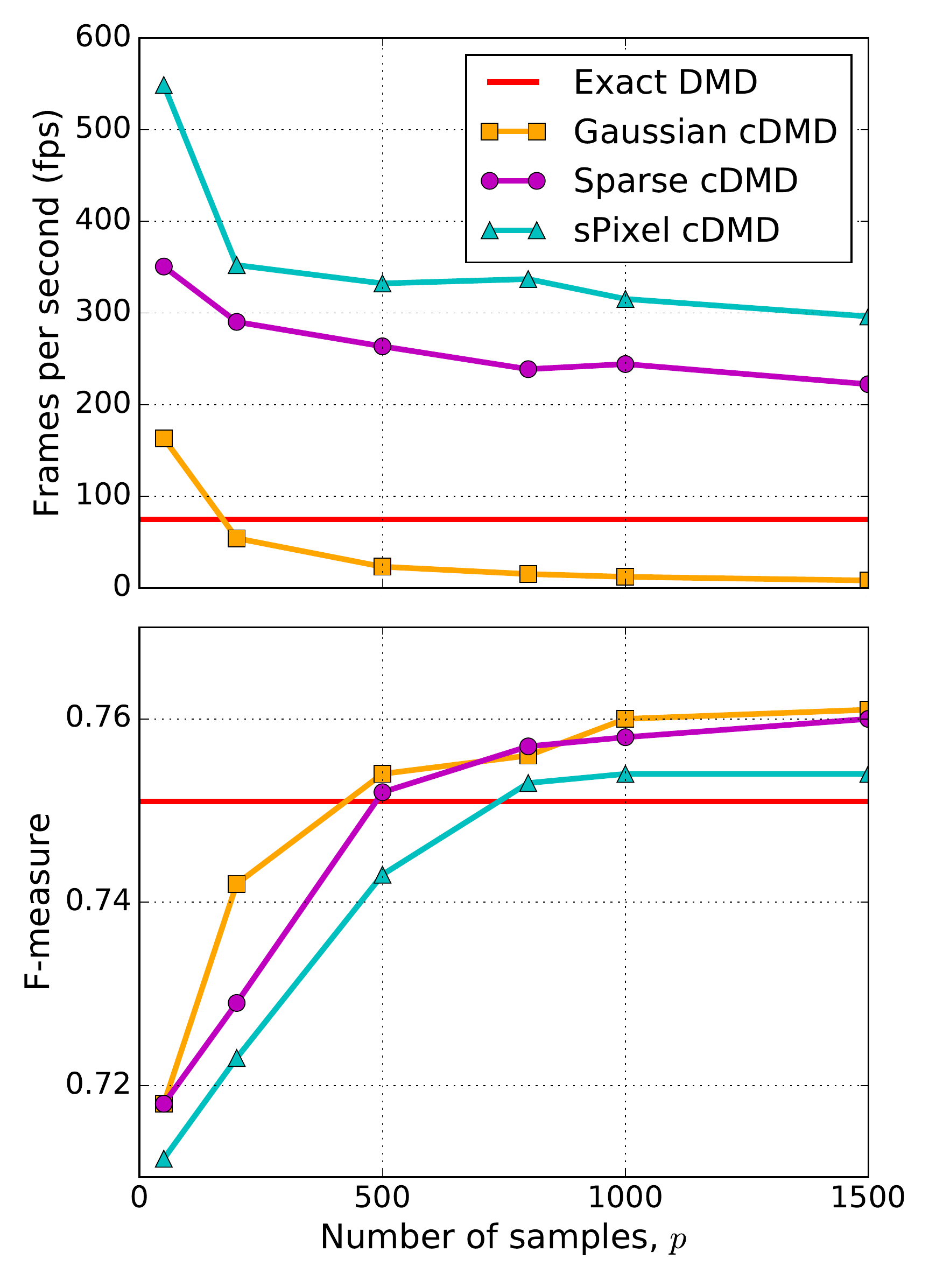}
	\caption{Algorithms runtime (excluding computation of the foreground mask) and accuracy for a varying number of samples $p$. Here a $720\times 480$ video sequence with 200 frames is used.}
	\label{fig:cDMDperformance}
\end{figure}

Figure~\ref{fig:cDMDperformance} investigates the performance of the different measurement matrices in more detail. Therefor, the fps rate and the F-measure is plotted for a varying number of samples $p$. 
Gaussian measurements achieves the best accuracy in terms of the F-measure, but the computational costs become increasingly expensive. Single pixel measurements (sPixel) is the most computationally efficient method. The primary advantages of single pixel measurements are the memory efficiency and the simple implementation. Sparse sensing matrices offer the best trade-off between computational time and accuracy, but require access to sparse matrix packages.

\begin{figure}[htp]
	\centering
	\DeclareGraphicsExtensions{.pdf}
	\includegraphics[width=0.5\textwidth]{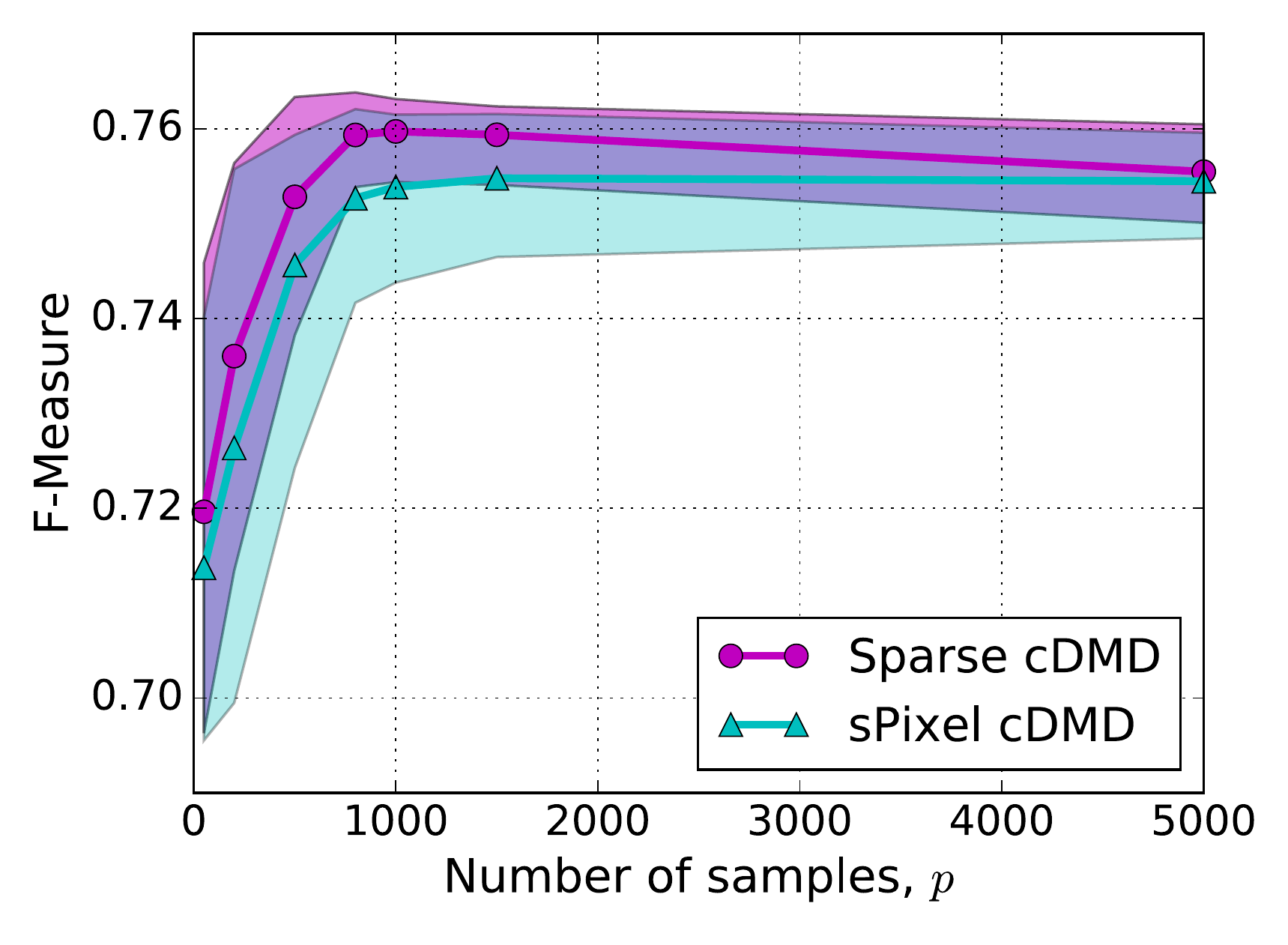}
	\caption{Bootstrap $95\%$-confidence intervals of the F-measure computed using both sparse and single pixel measurements.}
	\label{fig:Fvariation}
\end{figure}

It is important to stress that randomized sensing matrices cause random fluctuations influencing the background model quality, illustrated in Figure~\ref{fig:Fvariation}. The bootstrap confidence intervals show that sparse measurements have lower dispersion than single pixel measurements. This is, because single pixel measurements discard more information than sparse and Gaussian sensing matrices. 

\begin{table*}[!htpb]
	\centering
	\scalebox{0.85}{
		\begin{tabular}{ l l c c c c c c c c c c c} 
			\hline 			\hline
			& \multicolumn{1}{l}{Measure}
			& \multicolumn{9}{c}{BMC real videos}
			& \multicolumn{1}{c}{Average}
			\\
			\cmidrule(r){3-11}
			&  				& 001   & 002   & 003 & 004  & 005  & 006   & 007 & 008 & 009 \\
			\hline
			
			\multirow{3}{*}{\rotatebox[origin=c]{0}{ \parbox{3.5cm}{RSL \\ De La Torre et al. \cite{RPCA2} }  }} 
			& Recall 		& 0.800 & 0.689 & 0.840 & 0.872 & 0.861 & 0.823 & 0.658 & 0.589 & 0.690 & -
			\\
			& Precision    	& 0.732 & 0.808 & 0.804 & 0.585 & 0.598 & 0.713 & 0.636 & 0.526 & 0.625 & -
			\\
			& F-Measure 	& \textbf{0.765} & \textbf{0.744} & \textbf{0.821} & 0.700 & \textbf{0.706} & \textbf{0.764} & 0.647 & 0.556 & 0.656 & 0.707
			\\ \hline			
			
			\multirow{3}{*}{\rotatebox[origin=c]{0}{ \parbox{3.5cm}{LSADM \\ Goldfarb et al. \cite{Goldfarb} }     }} 
			& Recall 		& 0.693 & 0.535 & 0.784 & 0.721 & 0.643 & 0.656 & 0.449 & 0.621 & 0.701 & -
			\\
			& Precision    	& 0.511 & 0.724 & 0.802 & 0.729 & 0.475 & 0.655 & 0.693 & 0.633 & 0.809 & -
			\\
			& F-Measure 	& 0.591 & 0.618 & 0.793 & 0.725 & 0.549 & 0.656 & 0.551 & \textbf{0.627} & \textbf{0.752} & 0.650
			\\ \hline
			
			\multirow{3}{*}{\rotatebox[origin=c]{0}{ \parbox{3.5cm}{GoDec \\ Zhou and Tao \cite{Zhou11godec}}   }} 
			& Recall 		& 0.684 & 0.552 & 0.761 & 0.709 & 0.621 & 0.670 & 0.465 & 0.598 & 0.700  & -
			\\
			& Precision    	& 0.444 & 0.682 & 0.808 & 0.728 & 0.462 & 0.636 & 0.626 & 0.601 & 0.747  & -
			\\
			& F-Measure 	& 0.544 & 0.611 & 0.784 & 0.718 & 0.533 & 0.653 & 0.536 & 0.600 & 0.723 & 0.632 
			\\ \hline								
			
			\multirow{3}{*}{\rotatebox[origin=c]{0}{ \parbox{3.5cm}{ cDMD \\ } }} 
			& Recall 		& 0.552 & 0.697 & 0.778 & 0.693 & 0.611 & 0.700 & 0.720 & 0.515 & 0.566  & -
			\\
			& Precision    	& 0.581 & 0.675 & 0.773 & 0.770 & 0.541 & 0.602 & 0.823 & 0.510 & 0.574  & -
			\\
			& F-Measure 	& 0.566 & 0.686 & 0.776 & \textbf{0.730} & 0.574 & 0.647 & \textbf{0.768} & 0.512 & 0.570  & 0.648
			\\ 			
			
			\hline \hline
		\end{tabular}
	}
	\caption{Evaluation results of nine real videos from the BMC dataset. For comparison, the results of three other leading robust PCA algorihtms are presented, adapted from \cite{bouwmans2015decomp}.}
	\label{Tab:evaluationBMC}
\end{table*}

%


\section{Conclusion and Outlook}
\label{sec:conclusion}
We have introduced the compressed dynamic mode decomposition as a novel algorithm for video background modeling. Although many techniques have been developed in the last decade and a half to accomplish this task, significant challenges remain for the computer vision community when fast processing of high-definition video is required. Indeed, real-time HD video analysis remains one of the grand challenges of the field. Our cDMD method provides compelling evidence that it is a viable candidate for meeting this grand challenge, even on standard CPU computing platforms. The frame rate per second is highly competitive compared to other stat-of-the-art algorithms, e.g. Gaussian mixture-based algorithms. Compared to current robust principal component analysis based algorithm the increase in speed is even more substantial. In particular, the GPU accelerated implementation substantially improves the computational time. 

Despite the significant computational savings, the cDMD remains competitive with other leading
algorithms in the quality of the decomposition itself. Our results show, that for both standard and challenging environments, the cDMD's object detection accuracy in terms of the F-measure is competitive to leading RPCA based algorithms~\cite{bouwmans2015decomp}. Though, the algorithm cannot compete, in terms of the F-measure,  with highly specialized algorithms, e.g. optimized Gaussian mixture-based algorithms for background modeling~\cite{Sobralreview}. The main difficulties arise when video feeds are heavily crowded or dominated by non-periodic dynamic background objects. Overall, the trade-off between speed and accuracy of compressed DMD is compelling.   

Future work will aim to improve the background subtraction quality as well as to integrate a number of innovative techniques. One technique that is particularly useful for object tracking is the multi-resolution DMD~\cite{kutzMRDMD}. This algorithm has been shown to be a potential method for target tracking applications. Thus one can envision the integration of multi-resolution ideas with cDMD, i.e. a multi-resolution compressed DMD method, in order to separate the foreground video into different dynamic targets when necessary.

\begin{acknowledgements}
We would like to express our gratitude to E. R. Davies, K. Manohar and the three anonymous reviewers for many helpful comments on an earlier version of this paper. 

JNK acknowledges support from Air Force Office of Scientific Research (FA9500-15-C-0039).
SLB acknowledges support from the Department of Energy under award DE-EE0006785.  
NBE acknowledges support from the UK Engineering and Physical Sciences Research Council (EP/L505079/1).
\end{acknowledgements}

\begin{appendix}
\section{Notation}\label{app:notation}

\begin{tabbing}
	XXXXXXXXXX \= \kill
	\textbf{Scalars} \\
	$k$ \> Number of modes (target-rank) \\	
	$p$ \> Number of samples (measurements) \\
	$s$ \> Number of sparse samples \\	
	$K$ \> Number of non-zero amplitudes \\		
	$n$ \> Number of pixels per video frame \\
	$m$ \> Number of video frames \\
	$\lambda$ \> Eigenvalue \\
	$\omega$ \> Continuous-time eigenvalue \\[5pt]	
	\textbf{Vectors}\\
	${\bf x} \in \mathbb{R}^{n}$ \> Flattened video frame \\  
	${\bf y} \in \mathbb{R}^{p}$ \> Compressed video frame \\
	${\bf \phi} \in \mathbb{R}^{n}$ \> DMD mode \\
	${\bf b} \in \mathbb{R}^{k}$ \> Amplitudes \\
	${\bf \bbeta} \in \mathbb{R}^{k}$ \> Sparsity-constrained amplitudes \\[5pt]					
	\textbf{Matrices}\\
	${\bf X,X'} \in \mathbb{R}^{n \times m-1}$ \> Left and right snapshot sequence \\
	${\bf Y,Y'} \in \mathbb{R}^{p \times m-1}$ \> Compressed left/right snapshot sequence \\			
	${\bf C} \in \mathbb{R}^{p \times n}$ \> Measurement matrix \\
	${\bf A} \in \mathbb{R}^{n \times n}$ \> Linear map \\	
	${\bf \tilde{A}} \in \mathbb{R}^{k \times k}$ \> Rank-reduced linear map \\		
	${\bf \Phi} \in \mathbb{R}^{n \times k}$ \> DMD modes \\
	${\bf \Phi_Y} \in \mathbb{R}^{p \times k}$ \> Compressed DMD modes \\
	${\bf W, W_Y} \in \mathbb{R}^{k \times k}$ \> Rank-reduced eigenvectors \\	
	${\bf \Lambda, \Lambda_Y} \in \mathbb{R}^{k \times k}$ \> Rank-reduced eigenvalues (diagonal matrix) \\
	${\bf B} \in \mathbb{R}^{k \times k}$ \> Amplitudes (diagonal matrix) \\	
	${\bf \mathcal{V}} \in \mathbb{R}^{k \times m}$ \> Vandermonde matrix \\
	${\bf U_Y} \in \mathbb{R}^{p \times k}$ \> Truncated compressed left singular vectors \\	
	${\bf V_Y} \in \mathbb{R}^{k \times m-1}$ \> Truncated compressed right singular vectors \\
	${\bf S_Y} \in \mathbb{R}^{k \times k}$ \> Truncated compressed singular values \\						
\end{tabbing}
\end{appendix}

\bibliographystyle{elsarticle-num}
\bibliography{mrdmd,csdmd,BackSubtractbib,rdmdbib}   
\end{document}